\documentclass[twoside,11pt]{article}

\usepackage[preprint]{jmlr2e}
\usepackage{amsmath}
\usepackage{amsfonts}
\usepackage{mathrsfs}
\usepackage{algorithm}
\usepackage{algpseudocode}
\usepackage{graphicx}
\usepackage{booktabs}
\usepackage[figuresright]{rotating}
\usepackage{lastpage}
\usepackage{tikz}
\usetikzlibrary{arrows.meta,shapes.geometric}
\hypersetup{hidelinks,hypertexnames=false}

\definecolor{coverblue}{RGB}{0,114,178}
\definecolor{coverorange}{RGB}{213,94,0}
\definecolor{covergreen}{RGB}{0,132,104}
\definecolor{figHeaderGreen}{RGB}{230,239,225}
\definecolor{figHeaderOrange}{RGB}{255,240,213}
\definecolor{figHeaderRed}{RGB}{254,222,222}
\definecolor{figBlueDark}{RGB}{31,85,165}
\definecolor{figBlueMid}{RGB}{102,152,216}
\definecolor{figBlueLight}{RGB}{186,209,238}
\definecolor{figBluePale}{RGB}{238,244,250}
\definecolor{figOrangeDark}{RGB}{244,123,6}
\definecolor{figOrangeMid}{RGB}{254,188,125}
\definecolor{figOrangeLight}{RGB}{253,222,190}
\definecolor{figOrangePale}{RGB}{252,246,240}
\definecolor{figGreenDark}{RGB}{53,117,37}
\definecolor{figGreenMid}{RGB}{153,192,136}
\definecolor{figGreenLight}{RGB}{210,226,204}
\definecolor{figGreenPale}{RGB}{239,244,237}
\definecolor{figCellGray}{RGB}{245,245,245}
\definecolor{figBlueNode}{RGB}{2,80,180}
\definecolor{figOrangeNode}{RGB}{237,113,19}
\definecolor{figGreenArrow}{RGB}{86,143,70}

\newcommand{\change}[1]{#1}
\newenvironment{revision}{}{}
\allowdisplaybreaks

\newcommand{\COVER}{\textnormal{COVER-MTL}}
\newcommand{\Range}{\operatorname{Range}}
\newcommand{\Null}{\operatorname{Null}}

\newtheorem{lemma}{Lemma}

\newtheorem{proposition}{Proposition}

\newtheorem{corollary}{Corollary}
\newtheorem{assumption}{Assumption}
\newenvironment{proofof}[1]
{\par\noindent{\bf Proof of #1\ }}
{\hfill\BlackBox\\[2mm]}

\ShortHeadings{Covariate-Overlap Regularized MTL}{Sui, Xu, Bai and Qu}
\firstpageno{1}

\begin{document}

\title{Multi-Task Learning with Covariate-Overlap Regularization}

\author{%
\name Yang Sui \email yangsui@ucla.edu \\
\addr Department of Statistics and Data Science\\
University of California, Los Angeles\\
Los Angeles, CA 90095, USA
\AND
\name Qi Xu \email qixu@umn.edu \\
\addr School of Statistics\\
University of Minnesota\\
Minneapolis, MN 55455, USA
\AND
\name Yang Bai \email statbyang@mail.shufe.edu.cn \\
\addr School of Statistics and Data Science\\
Shanghai University of Finance and Economics\\
Shanghai 200433, China
\AND
\name Annie Qu \email aqu2@ucsb.edu \\
\addr Department of Statistics and Applied Probability\\
University of California, Santa Barbara\\
Santa Barbara, CA 93106, USA
}

\maketitle
\thispagestyle{plain}

\begin{abstract}%
Multi-task learning improves data efficiency by sharing information across related tasks, but indiscriminate sharing can be harmful when their covariate distributions and response relationships differ. We propose COVariate-ovERlap regularized multi-task learning (\COVER) to address covariate and posterior heterogeneity. The model combines a common component function with a shared neural representation and low-dimensional task-specific coefficients. Taskwise second-moment matrices summarize covariate heterogeneity and determine the strength of coefficient integration in each representation direction. We derive a covariate-overlap penalty by minimizing the total squared change in two task predictors when their coefficients are replaced by one auxiliary coefficient. An equivalent auxiliary formulation supports end-to-end training without matrix inversion. An exact fixed-representation bias--variance decomposition quantifies how covariate overlap controls variance reduction and how posterior heterogeneity determines shrinkage bias. Global and localized end-to-end oracle inequalities account for jointly learning the neural functions and estimating the overlap matrices from the same observations. We give explicit neural-network rates and sharpen the stochastic prediction term when the regularized oracle risk and overlap-estimation error are small. Simulations across diverse heterogeneity settings show competitive performance against deep-learning and statistical data-integration methods, with the largest gains under joint heterogeneity. In a GTEx central-nervous-system analysis, \COVER{} achieves the lowest response-averaged prediction error among the compared methods and reveals tissue-pair integration patterns.
\end{abstract}


\section{Introduction}\label{sec:introduction}

Many scientific prediction problems comprise related data sets that are individually too small to support accurate models. Multi-task learning (MTL) improves data efficiency by learning these tasks jointly and has been used in biomedical research, computer vision, and natural language processing \citep{wang2019cross,gao2019nddr,sanh2019hierarchical,zhang2018overview}. Its benefit nevertheless depends on a central statistical question: what should be integrated across tasks, and in which directions do the data support integration?

Two forms of heterogeneity make this question difficult. First, tasks may have different covariate distributions, a setting we call covariate heterogeneity. This setting is also commonly referred to as covariate shift \citep{shimodaira2000improving}. Such differences arise when patient populations, molecular profiles, or measurement distributions vary across disease subtypes, tissues, or study sites \citep{gtex2020atlas}. Second, the conditional distribution of the response given the covariates may vary across tasks, which we call posterior heterogeneity. Changes in the conditional distribution are also called posterior drift in transfer learning \citep{maity2024linear} and are related to the broader concept-drift setting of \citet{widmer1996learning}. For example, the same molecular profile may have different associations with an outcome across tissues or cancer types. Ignoring posterior heterogeneity oversmooths scientifically meaningful task differences. Ignoring covariate heterogeneity is also consequential, because a task-specific relationship should not be strongly transferred along representation directions that are well supported in one task but scarcely observed in another.

Complex covariate--response relationships compound these difficulties. Genomic and imaging signals can involve nonlinear effects, high-order interactions, and regulatory or spatial structure poorly approximated by linear models \citep{litjens2017survey}. Neural networks can learn such relationships and lower-dimensional representations, but do not by themselves resolve the integration problem: a common network can obscure task-specific effects, whereas separate networks can be statistically and computationally inefficient. A useful nonlinear MTL method should therefore explain how both forms of heterogeneity affect integration.

Statistical data-integration methods can be viewed in two broad categories. The first addresses covariate heterogeneity through importance weighting or related corrections to the empirical risk \citep{shimodaira2000improving,ma2023optimally}. In particular, \citet{shimodaira2000improving} showed how weighting the likelihood can correct predictive inference under covariate shift. The second regulates information borrowing when response relationships differ, through parameter fusion, adaptive multi-task constraints, posterior-drift adjustment, or estimator combination \citep{tang2016fused,duan2023adaptive,maity2024linear,li2024combining}. For structured imaging coefficients, \citet{sui2025pair} used a partially shared fusion penalty to identify components that can be integrated across diagnostic groups. These two categories provide useful tools for their respective forms of heterogeneity, but they typically treat the other form as secondary. Many also rely on linear or otherwise structured models, and conventional coefficient-fusion penalties do not use the taskwise covariate distributions to decide where two coefficients should be pooled.

Network-based MTL offers a complementary set of approaches. Existing methods can be grouped roughly into loss or gradient balancing, architectural parameter sharing, and representation-based integration. The first group modifies task losses or gradients to stabilize joint optimization \citep{kendall2018multi,yu2020gradient}; the second includes hard and soft parameter sharing, which respectively share network layers or regularize task-specific parameters \citep{caruana1997multitask,zhang2018overview}; and the third learns shared, task-specific, or selectively retrieved representations \citep{tripuraneni2020theory,watkins2024optimistic,sui2026multitask,xu2026representation}. These methods greatly expand the class of relationships that can be learned. Yet the amount of sharing is usually determined by a common architecture, a coordinatewise parameter penalty, or a selected collection of representers. By comparison, taskwise covariate distributions are not generally used to determine both the strength and the representation directions of coefficient integration.

\begin{revision}
We develop COVariate-ovERlap regularized multi-task learning (\COVER) to address this gap. For task \(t\), we write the conditional mean as \(f_t(x)=g(x)+z(x)^\top\beta_t\), where \(g\) is a common component function, \(z\) is a shared representation, and \(\beta_t\) is a low-dimensional task-specific coefficient. A centering constraint makes \(g\) the task-balanced common component function and expresses every pairwise conditional-mean difference through \(z(x)^\top(\beta_t-\beta_s)\). This decomposition is related to hard parameter sharing and multi-task representation learning \citep{caruana1997multitask,tripuraneni2020theory}, but its role here is to provide a common function space in which coefficient differences can be integrated. Posterior heterogeneity is represented by \(\{\beta_t\}_{t=1}^T\), whereas covariate heterogeneity enters through \(\Sigma_t=\mathbb E_t\{z(X)z(X)^\top\}\), the task-\(t\) second-moment matrix of the shared representation.
\end{revision}

\begin{revision}
To account for covariate heterogeneity when integrating task-specific coefficients, we propose a novel covariate-overlap penalty. If two tasks have the same second-moment matrix, \(\Sigma_t=\Sigma_s\), their coefficient difference can be penalized using the common second-moment matrix. When \(\Sigma_t\ne\Sigma_s\), however, using either matrix is asymmetric, while averaging them can over-penalize directions represented mainly in only one task. We instead define the pairwise pooling cost as the minimum total squared change in the two task predictors, measured under their respective covariate distributions, when \(\beta_t\) and \(\beta_s\) are replaced by one auxiliary coefficient. This minimization yields an overlap matrix that assigns large weights to jointly supported representation directions and small weights to weakly overlapping directions. The resulting penalty therefore integrates \(\beta_t\) and \(\beta_s\) strongly only where both tasks support the corresponding prediction change. Under equal second moments, it reduces to a quadratic penalty weighted by the common second-moment matrix; as covariate overlap decreases, it gradually weakens integration.
\end{revision}

\begin{revision}
Our theoretical analysis explains how the two forms of heterogeneity govern data integration. As basic structural properties, \COVER{} does not create task-specific components from covariate heterogeneity alone when the common component function is correctly specified, and its fitted functions and penalty values are invariant to invertible coordinate changes within a fixed shared representation span. For a fixed representation, an exact direction-wise bias--variance decomposition makes their distinct roles explicit: the taskwise covariate distributions determine the overlap matrices and the integration strength along each representation direction, while coefficient differences encode conditional-mean heterogeneity and, together with these matrices, determine shrinkage bias. Thus, directions well supported by multiple tasks permit greater variance reduction, while shrinking a genuine conditional-mean difference incurs bias. We next establish an end-to-end risk bound for jointly learned neural functions that separates approximation and shrinkage errors, neural-class complexity, sampling fluctuation, and overlap-matrix estimation. Building on localized complexity arguments \citep{bartlett2005local}, we further obtain a faster stochastic order in low-risk regimes. Together, these results clarify when the covariate-overlap penalty improves prediction through data integration and what is required to estimate the integration geometry from the same observations. We also study its theoretical performance when transferring information to a new but related task.
\end{revision}

\begin{revision}
The proposed criterion also has a direct computational form. Introducing one auxiliary coefficient for each task pair converts the overlap penalty into a sum of empirical squared prediction differences. Minimizing over the auxiliary coefficients recovers exactly the matrix-weighted criterion, but training the auxiliary formulation requires no matrix inversion. The common component function, shared representation, task-specific coefficients, and auxiliary coefficients can consequently be optimized in a single differentiable objective using standard backpropagation. We evaluate \COVER{} through simulations under a variety of settings, including homogeneous data and data with covariate-only, posterior-only, or joint heterogeneity, among others. We compare it with a range of advanced deep-learning methods and statistical multi-task penalties. \COVER{} is most beneficial when covariate and posterior heterogeneity occur together, and the mechanism experiments confirm that its integration strength changes with the learned overlap geometry. It also attains the lowest excess prediction error throughout the reported task- and dimension-scaling experiments. In the Genotype--Tissue Expression central-nervous-system analysis, all learned methods improve substantially over mean prediction, and \COVER{} has the best response-averaged performance across the two gene-expression outcomes. The learned overlap matrices additionally reveal tissue-pair integration patterns that agree qualitatively with established transcriptomic structure.
\end{revision}

The remainder of the paper is organized as follows. Section~\ref{sec:method} introduces the problem setup and develops the \COVER{} model and covariate-overlap penalty. Section~\ref{sec:computation} presents the auxiliary formulation and training procedure. Section~\ref{sec:theory} establishes the theoretical properties, and Section~\ref{sec:experiments} reports the simulation and GTEx analyses. Section~\ref{sec:discussion} discusses limitations and future directions. Appendices~\ref{app:additional-experiments}--\ref{app:main-proofs} provide additional experiments, the multiway-pooling extension, supplementary theory, new-task transfer, and proofs, respectively.

\section{The \COVER{} Method}\label{sec:method}

This section introduces \COVER, whose sharing structure combines a common component function and a shared representation with task-specific coefficients that represent conditional-mean heterogeneity. We also propose a covariate-overlap penalty that uses covariate heterogeneity to determine how these coefficients are integrated.

\subsection{Preliminaries}\label{sec:setting}

We first introduce the problem setup and essential notation. For any positive integer \(m\), let \([m]=\{1,\ldots,m\}\), and use \([T]\) to index the tasks. For task \(t\in[T]\), let \(\mathbb P_t\) denote the joint population distribution of \((X,Y)\in\mathbb R^p\times\mathbb R\). The observed data are \(\mathcal D_t=\{(X_{ti},Y_{ti})\}_{i=1}^{n_t}\), where \((X_{ti},Y_{ti})\overset{\mathrm{iid}}{\sim}\mathbb P_t\). Uppercase letters denote random variables and observations, while lowercase \(x\) is reserved for a fixed covariate argument. The empirical method permits unequal \(n_t\), whereas Section~\ref{sec:theory} sets \(n_t=n\) to simplify the main theoretical expressions. The covariate distribution and conditional distribution of task \(t\) are denoted by \(\mathbb P_t^X\) and \(\mathbb P_t^{Y\mid X}\), respectively. We write \(\mathbb E_t\) for expectation under \(\mathbb P_t\) and \(\mathbb E_t(\,\cdot\mid X=x)\) for conditional expectation under \(\mathbb P_t^{Y\mid X}\). An expectation without a task subscript is taken over the random quantities specified in the surrounding statement.
Two tasks \(t,s\in[T]\) with \(t\ne s\) exhibit covariate heterogeneity when \(\mathbb P_t^X\ne\mathbb P_s^X\) and posterior heterogeneity when \(\mathbb P_t^{Y\mid X}\ne\mathbb P_s^{Y\mid X}\). Under squared loss, the task-specific estimand is the conditional mean \(f_t^\ast(x)=\mathbb E_t(Y\mid X=x)\). In this paper, we target conditional-mean heterogeneity in \(\{f_t^\ast\}_{t=1}^T\).

The shared representation dimension is \(d\), and \(I_d\) denotes the \(d\times d\) identity matrix. For \(u\in\mathbb R^d\) and a positive semidefinite matrix \(A\succeq0\), write \(\|u\|_A^2=u^\top A u\). The notation \(A\preceq B\) means that \(B-A\) is positive semidefinite, and \(A^\dagger\) denotes the Moore--Penrose inverse. We use \(\Range(A)=\{Av:v\in\mathbb R^d\}\) for the column space of \(A\) and \(\Null(A)=\{v\in\mathbb R^d:Av=0\}\) for its null space. For a symmetric positive semidefinite matrix, these are respectively the spans of the eigenvectors with positive and zero eigenvalues, and \(\Range(A)=\Null(A)^\perp\).

\subsection{Common Component Function and Task-Specific Coefficients}\label{sec:model}

We model the task-\(t\) conditional mean by
\begin{revision}
\begin{equation}\label{eq:model}
    f_t(x)=g(x)+z(x)^\top\beta_t,
    \qquad
    \sum_{t=1}^T\beta_t=0.
\end{equation}
\end{revision}
Here \(g:\mathbb R^p\to\mathbb R\) is the common component function, \(z:\mathbb R^p\to\mathbb R^d\) is the shared representation, and \(\beta_t\in\mathbb R^d\) is the task-specific coefficient. The term \(z(x)^\top\beta_t\) is the task-specific component function. In the GTEx analysis of Section~\ref{sec:gtex}, for example, \(g(x)\) can capture associations between the module-gene profile and the response that persist across central-nervous-system tissues, whereas \(z(x)^\top\beta_t\) captures tissue-specific departures when predicting the expression of \textit{JAM2} or \textit{SH2D2A} \citep{gtex2020atlas}.

\begin{revision}
The decomposition in \eqref{eq:model} is related to the use of shared functions and task-specific output coefficients in hard parameter sharing and multi-task representation learning \citep{caruana1997multitask,zhang2018overview,tripuraneni2020theory,watkins2024optimistic}. We impose the centering constraint in \eqref{eq:model}, which fixes the decomposition through \(g(x)=T^{-1}\sum_{t=1}^T f_t(x)\) and gives \(f_t(x)-f_s(x)=z(x)^\top(\beta_t-\beta_s)\). We assess coefficient integration through its effect on the task predictors: the taskwise covariate distributions determine how strongly, and along which representation directions, each pair of task-specific coefficients \((\beta_t,\beta_s)\) is integrated. Such covariate-aware integration has not been systematically studied.
\end{revision}

\subsection{Covariate-Overlap Penalty}\label{sec:overlap}

The most direct way to perform data integration is to penalize \(\|\beta_t-\beta_s\|_2^2\). This Euclidean penalty is reasonable for a fixed, standardized representation, but it depends on the coordinates and scale of a jointly optimized \(z\). For example, replacing \(z\) by \(cz\) and every \(\beta_t\) by \(\beta_t/c\) leaves all fitted functions unchanged while dividing the Euclidean penalty by \(c^2\). We therefore measure changes in the task predictors rather than distances between coefficient coordinates.

For a fixed shared representation, define the task-specific representation second-moment matrix \(\Sigma_t=\mathbb E_t\{z(X)z(X)^\top\}\). It is a functional of the covariate distribution \(\mathbb P_t^X\) and satisfies \(\|v\|_{\Sigma_t}^2=\mathbb E_t\{z(X)^\top v\}^2\). In particular, for any \(a\in\mathbb R^d\),
\[
\|\beta_t-a\|_{\Sigma_t}^2
=\mathbb E_t\!\left[\{z(X)^\top\beta_t-z(X)^\top a\}^2\right].
\]
Holding \(g\) and \(z\) fixed, this measures the expected squared change in task \(t\)'s predictions when \(\beta_t\) is replaced by \(a\). The identity also holds for singular \(\Sigma_t\), when the quadratic form is a seminorm: \(v\in\Null(\Sigma_t)\) exactly when \(z(X)^\top v=0\) almost surely under \(\mathbb P_t^X\), so \(\Range(\Sigma_t)\) describes the representation directions present in task \(t\). When \(\Sigma_t=\Sigma_s=\Sigma\), taking \(a=\beta_s\) gives \(\|\beta_t-\beta_s\|_\Sigma^2=\mathbb E_t\{f_t(X)-f_s(X)\}^2\), which measures the prediction difference between the two tasks. This penalty reduces to the Euclidean penalty after whitening and is invariant to invertible reparameterizations of the shared representation.

When \(\Sigma_t\ne\Sigma_s\), however, there is no common prediction norm. Using either task's matrix is asymmetric. Averaging the two taskwise norms is symmetric, but it can over-regularize a direction represented mainly in one task: the weight \((\Sigma_t+\Sigma_s)/2\) reflects the union rather than the shared representation subspace. In a one-dimensional example with \(\Sigma_t=1\) and \(\Sigma_s=0.01\), the Average-Moment weight is \(0.505\), even though task \(s\) contains almost no information in that direction. Strongly shrinking the two coefficients there can transfer a relationship that the second task cannot validate.

For unequal second-moment matrices, we instead ask how to replace both coefficients by one auxiliary coefficient \(a\) while changing the two task predictors as little as possible. Summing their expected squared changes and minimizing over \(a\) gives the pairwise pooling cost
\begin{equation}\label{eq:pooling-cost}
\mathcal C_{ts}(\beta_t,\beta_s)
=\min_{a\in\mathbb R^d}
\left\{\|\beta_t-a\|_{\Sigma_t}^2+\|\beta_s-a\|_{\Sigma_s}^2\right\}.
\end{equation}
This criterion treats the tasks symmetrically while evaluating each change under the corresponding covariate distribution. The following proposition profiles out \(a\).

\begin{revision}
\begin{proposition}[Pairwise pooling identity]\label{prop:pooling}
For any \(\Sigma_t,\Sigma_s\succeq0\),
\begin{equation}\label{eq:overlap}
\mathcal C_{ts}(\beta_t,\beta_s)
=\frac12\|\beta_t-\beta_s\|_{\Omega_{ts}}^2,
\qquad
\Omega_{ts}=2\Sigma_t(\Sigma_t+\Sigma_s)^\dagger\Sigma_s.
\end{equation}
Moreover, \(\Omega_{ts}\) is symmetric and positive semidefinite, with \(\Range(\Omega_{ts})=\Range(\Sigma_t)\cap\Range(\Sigma_s)\).
\end{proposition}
\end{revision}

When both second-moment matrices are positive definite, the minimizing auxiliary coefficient is \(a_{ts}^{\ast}=(\Sigma_t+\Sigma_s)^{-1}(\Sigma_t\beta_t+\Sigma_s\beta_s)\), and \(\Omega_{ts}=2(\Sigma_t^{-1}+\Sigma_s^{-1})^{-1}\). In particular, \(\Omega_{ts}=\Sigma\) when \(\Sigma_t=\Sigma_s=\Sigma\), recovering the usual prediction-norm penalty. The range of \(\Omega_{ts}\) is the shared representation subspace. Thus, coefficient differences are not penalized in directions absent from either task. Directions weakly represented in one task receive smaller penalty weights, even when both estimated matrices are full rank.

The resulting covariate-overlap penalty can be compared directly with the Average-Moment penalty.

\begin{corollary}[Average-Moment Comparison]\label{cor:penalty-comparison}
For any \(\Sigma_t,\Sigma_s\succeq0\),
\begin{equation}\label{eq:penalty-comparison}
\frac{\Sigma_t+\Sigma_s}{2}-\Omega_{ts}
=\frac12(\Sigma_t-\Sigma_s)(\Sigma_t+\Sigma_s)^\dagger
(\Sigma_t-\Sigma_s)\succeq0.
\end{equation}
Consequently, \(\|\beta_t-\beta_s\|_{\Omega_{ts}}^2\) is no larger than the average of \(\|\beta_t-\beta_s\|_{\Sigma_t}^2\) and \(\|\beta_t-\beta_s\|_{\Sigma_s}^2\), with equality of the two weighting matrices when \(\Sigma_t=\Sigma_s\).
If, moreover, \(\Sigma_t=\Sigma\) for every task and \(\sum_{t=1}^T\beta_t=0\), then \(\Omega_{ts}=\Sigma\) and \(\sum_{1\le t<s\le T}\|\beta_t-\beta_s\|_\Sigma^2=T\sum_{t=1}^T\|\beta_t\|_\Sigma^2\). Hence \COVER{} reduces to prediction-norm ridge shrinkage of the centered task-specific coefficients.
\end{corollary}

\begin{revision}
Thus, the covariate-overlap penalty equals the Average-Moment penalty minus a positive semidefinite correction determined by covariate heterogeneity. In the one-dimensional example, the weight falls from \(0.505\) to approximately \(0.0198\). This reduces over-pooling bias when genuine task differences occur outside the shared representation subspace, but may retain more variance when task-specific coefficients are equal. When all tasks share the same second-moment matrix, the correction vanishes, yielding ridge shrinkage weighted by that matrix. This reduction also explains the normalization by \(T(T-1)\), which keeps the effective regularization strength comparable as the number of tasks changes.
\end{revision}

The proofs of Proposition~\ref{prop:pooling} and Corollary~\ref{cor:penalty-comparison} are provided in Appendix~\ref{app:method-proofs}. Together, these results give the data-integration rule: task-specific coefficients are pooled only along the shared representation subspace, while ordinary global pooling is recovered when the second-moment matrices are equal. Appendix~\ref{app:multiway-pooling} records an extension to pooling three or more tasks and explains why the main method uses pairs. Figure~\ref{fig:overlap-geometry} illustrates how increasing covariate heterogeneity weakens the covariate-overlap penalty.

\begin{figure}[t]
\centering
\resizebox{0.95\textwidth}{!}{%
\begin{tikzpicture}[
    x=1cm,
    y=1cm,
    paneltitle/.style={font=\scriptsize\sffamily\bfseries,align=center},
    rowlabel/.style={font=\scriptsize\sffamily\bfseries,align=center},
    smalllabel/.style={font=\scriptsize\sffamily,align=center},
    axis/.style={-{Latex[length=1.5mm]},draw=black!72,line width=0.55pt},
    inward/.style={-{Latex[length=2.1mm,width=1.55mm]},draw=figGreenArrow,line width=1.15pt}
]
\path[fill=figHeaderGreen,rounded corners=4pt] (-2.25,4.65) rectangle (2.25,4.08);
\path[fill=figHeaderOrange,rounded corners=4pt] (2.75,4.65) rectangle (7.25,4.08);
\path[fill=figHeaderRed,rounded corners=4pt] (7.75,4.65) rectangle (12.25,4.08);
\node[paneltitle] at (0,4.36) {No covariate heterogeneity};
\node[paneltitle] at (5,4.36) {Moderate covariate heterogeneity};
\node[paneltitle] at (10,4.36) {Strong covariate heterogeneity};
\draw[black!18,densely dashed,line width=0.5pt] (2.50,3.95)--(2.50,-4.25);
\draw[black!18,densely dashed,line width=0.5pt] (7.50,3.95)--(7.50,-4.25);
\draw[black!16,line width=0.45pt] (-2.25,1.22)--(12.25,1.22);
\draw[black!16,line width=0.45pt] (-2.25,-1.22)--(12.25,-1.22);

\begin{scope}[shift={(0,1.45)}]
  \draw[axis] (-1.70,0)--(1.82,0);
  \draw[axis] (-1.70,0)--(-1.70,2.28);
  \begin{scope}[rotate around={27:(0,1.02)}]
    \draw[draw=figBlueNode,fill=figBlueLight,fill opacity=.42,line width=.75pt]
      (0,1.02) ellipse[x radius=1.55,y radius=.58];
    \draw[draw=figOrangeNode,fill=figOrangeLight,fill opacity=.42,line width=.75pt]
      (.08,1.00) ellipse[x radius=1.46,y radius=.55];
  \end{scope}
  \node[font=\large,text=figBlueNode] at (-.92,1.90) {$\Sigma_t$};
  \node[font=\large,text=figOrangeNode] at (.97,.45) {$\Sigma_s$};
\end{scope}

\begin{scope}[shift={(5,1.45)}]
  \draw[axis] (-1.70,0)--(1.82,0);
  \draw[axis] (-1.70,0)--(-1.70,2.28);
  \begin{scope}[rotate around={72:(-.25,1.15)}]
    \draw[draw=figBlueNode,fill=figBlueLight,fill opacity=.45,line width=.75pt]
      (-.25,1.15) ellipse[x radius=1.08,y radius=.44];
  \end{scope}
  \begin{scope}[rotate around={8:(.28,.82)}]
    \draw[draw=figOrangeNode,fill=figOrangeLight,fill opacity=.45,line width=.75pt]
      (.28,.82) ellipse[x radius=1.32,y radius=.38];
  \end{scope}
  \draw[figGreenDark,densely dashed,line width=.8pt] (-1.00,.12)--(.93,2.10);
  \node[font=\large,text=figBlueNode] at (-1.28,2.02) {$\Sigma_t$};
  \node[font=\large,text=figOrangeNode,anchor=west] at (1.68,.75) {$\Sigma_s$};
\end{scope}

\begin{scope}[shift={(10,1.45)}]
  \draw[axis] (-1.70,0)--(1.82,0);
  \draw[axis] (-1.70,0)--(-1.70,2.28);
  \begin{scope}[rotate around={79:(-1.00,1.13)}]
    \draw[draw=figBlueNode,fill=figBlueLight,fill opacity=.45,line width=.75pt]
      (-1.00,1.13) ellipse[x radius=1.08,y radius=.38];
  \end{scope}
  \begin{scope}[rotate around={4:(.62,.78)}]
    \draw[draw=figOrangeNode,fill=figOrangeLight,fill opacity=.45,line width=.75pt]
      (.62,.78) ellipse[x radius=1.02,y radius=.34];
  \end{scope}
  \node[font=\large,text=figBlueNode] at (-1.36,2.07) {$\Sigma_t$};
  \node[font=\large,text=figOrangeNode,anchor=west] at (1.50,.25) {$\Sigma_s$};
\end{scope}

\node[font=\large,text=figGreenDark] at (-1.58,.92) {$\Omega_{ts}$};
\begin{scope}[rotate around={27:(0,.14)}]
  \draw[draw=figGreenDark,fill=figGreenLight,fill opacity=.72,line width=.8pt]
    (0,.14) ellipse[x radius=1.55,y radius=.50];
\end{scope}
\draw[figGreenDark,densely dashed,line width=.7pt] (-1.26,-.53)--(1.27,.81);
\node[smalllabel] at (0,-1.02) {Broad overlap};

\node[font=\large,text=figGreenDark] at (3.73,.92) {$\Omega_{ts}$};
\begin{scope}[rotate around={45:(5,.18)}]
  \draw[draw=figGreenDark,fill=figGreenLight,fill opacity=.78,line width=.8pt]
    (5,.18) ellipse[x radius=1.42,y radius=.18];
\end{scope}
\draw[figGreenDark,densely dashed,line width=.7pt] (4.08,-.74)--(5.92,1.10);
\node[smalllabel] at (5,-1.02) {Direction-selective overlap};

\node[font=\large,text=figGreenDark] at (10,.87) {$\Omega_{ts}\approx0$};
\draw[figGreenDark,line width=.75pt] (10,.04) circle (2.0pt);
\node[smalllabel] at (10,-1.02) {Negligible overlap};

\begin{scope}[shift={(0,-2.82)}]
  \draw[axis] (-1.70,-.72)--(1.82,-.72);
  \draw[axis] (-1.70,-.72)--(-1.70,1.30);
  \draw[-{Latex[length=2.0mm,width=1.4mm]},figBlueNode,line width=.9pt,shorten >=3pt]
    (-.85,.90)--(-.18,.39);
  \draw[-{Latex[length=2.0mm,width=1.4mm]},figOrangeNode,line width=.9pt,shorten >=3pt]
    (1.02,-.52)--(.36,-.03);
  \draw[figGreenDark,densely dashed,line width=.8pt] (-.18,.39)--(.09,.18);
  \draw[figGreenDark,densely dashed,line width=.8pt] (.36,-.03)--(.09,.18);
  \fill[figBlueNode] (-.85,.90) circle (2.6pt);
  \fill[figOrangeNode] (1.02,-.52) circle (2.6pt);
  \filldraw[fill=white,draw=figBlueNode,line width=.9pt] (-.18,.39) circle (3.0pt);
  \filldraw[fill=white,draw=figOrangeNode,line width=.9pt] (.36,-.03) circle (3.0pt);
  \fill[figGreenDark] (.09,.18) circle (2.5pt);
  \node[font=\large,text=figBlueNode,anchor=south west] at (-.72,1.01) {$\beta_t$};
  \node[font=\large,text=figOrangeNode,anchor=south west] at (1.10,-.32) {$\beta_s$};
  \node[font=\large,text=figGreenDark,anchor=south west] at (.16,.25) {$a_{ts}$};
\end{scope}
\node[rowlabel] at (0,-3.88) {Strong pooling};

\begin{scope}[shift={(5,-2.82)}]
  \draw[axis] (-1.70,-.72)--(1.82,-.72);
  \draw[axis] (-1.70,-.72)--(-1.70,1.30);
  \draw[-{Latex[length=2.0mm,width=1.4mm]},figBlueNode,line width=.9pt,shorten >=3pt]
    (-.85,.90)--(-.49,.62);
  \draw[-{Latex[length=2.0mm,width=1.4mm]},figOrangeNode,line width=.9pt,shorten >=3pt]
    (1.02,-.52)--(.68,-.26);
  \draw[figGreenDark,densely dashed,line width=.8pt] (-.49,.62)--(.095,.18);
  \draw[figGreenDark,densely dashed,line width=.8pt] (.68,-.26)--(.095,.18);
  \fill[figBlueNode] (-.85,.90) circle (2.6pt);
  \fill[figOrangeNode] (1.02,-.52) circle (2.6pt);
  \filldraw[fill=white,draw=figBlueNode,line width=.9pt] (-.49,.62) circle (3.0pt);
  \filldraw[fill=white,draw=figOrangeNode,line width=.9pt] (.68,-.26) circle (3.0pt);
  \fill[figGreenDark] (.095,.18) circle (2.5pt);
  \node[font=\large,text=figBlueNode,anchor=south west] at (-.72,1.01) {$\beta_t$};
  \node[font=\large,text=figOrangeNode,anchor=south west] at (1.10,-.32) {$\beta_s$};
  \node[font=\large,text=figGreenDark,anchor=north] at (.095,.10) {$a_{ts}$};
\end{scope}
\node[rowlabel] at (5,-3.88) {Direction-selective pooling};

\begin{scope}[shift={(10,-2.82)}]
  \draw[axis] (-1.70,-.72)--(1.82,-.72);
  \draw[axis] (-1.70,-.72)--(-1.70,1.30);
  \draw[-{Latex[length=1.2mm,width=.9mm]},figBlueNode,line width=.9pt,shorten >=3pt]
    (-.85,.90)--(-.62,.72);
  \draw[-{Latex[length=1.2mm,width=.9mm]},figOrangeNode,line width=.9pt,shorten >=3pt]
    (1.02,-.52)--(.79,-.34);
  \draw[figGreenDark,densely dashed,line width=.8pt] (-.62,.72)--(.095,.18);
  \draw[figGreenDark,densely dashed,line width=.8pt] (.79,-.34)--(.095,.18);
  \fill[figBlueNode] (-.85,.90) circle (2.6pt);
  \fill[figOrangeNode] (1.02,-.52) circle (2.6pt);
  \filldraw[fill=white,draw=figBlueNode,line width=.9pt] (-.62,.72) circle (3.0pt);
  \filldraw[fill=white,draw=figOrangeNode,line width=.9pt] (.79,-.34) circle (3.0pt);
  \fill[figGreenDark] (.095,.18) circle (2.5pt);
  \node[font=\large,text=figBlueNode,anchor=south west] at (-.72,1.01) {$\beta_t$};
  \node[font=\large,text=figOrangeNode,anchor=south west] at (1.10,-.32) {$\beta_s$};
  \node[font=\large,text=figGreenDark,anchor=north] at (.095,.10) {$a_{ts}$};
\end{scope}
\node[rowlabel] at (10,-3.88) {Little or no pooling};
\end{tikzpicture}
}
\caption{Covariate heterogeneity changes $\Omega_{ts}$ and the resulting pooling toward $a_{ts}$; filled and open circles denote coefficients before and after pooling.}
\label{fig:overlap-geometry}
\end{figure}

\subsection{Population Criterion and Empirical Estimator}\label{sec:estimator}

For a regularization parameter \(\lambda\ge0\), the population \COVER{} criterion is
\[
\begin{aligned}
\min_{g,z,\{\beta_t\}_{t=1}^T}\quad&
\frac1{2T}\sum_{t=1}^T\mathbb E_t\{Y-f_t(X)\}^2+
\frac{\lambda}{T(T-1)}
\sum_{1\le t<s\le T}\|\beta_t-\beta_s\|_{\Omega_{ts}}^2,\\
\text{subject to}\quad&\sum_{t=1}^T\beta_t=0.
\end{aligned}
\]
For a candidate representation \(z\), let \(\widehat\Sigma_t=n_t^{-1}\sum_{i=1}^{n_t}z(X_{ti})z(X_{ti})^\top\), and define \(\widehat\Omega_{ts}\) from \eqref{eq:overlap} by replacing the population second-moment matrices with their empirical counterparts. The empirical estimator is
\begin{equation}\label{eq:estimator}
\begin{aligned}
\min_{g,z,\{\beta_t\}_{t=1}^T}\quad&
\sum_{t=1}^T\frac1{2Tn_t}
\sum_{i=1}^{n_t}
\{Y_{ti}-f_t(X_{ti})\}^2+
\frac{\lambda}{T(T-1)}
\sum_{1\le t<s\le T}\|\beta_t-\beta_s\|_{\widehat\Omega_{ts}}^2,\\
\text{subject to}\quad&\sum_{t=1}^T\beta_t=0.
\end{aligned}
\end{equation}
Equation~\eqref{eq:estimator} defines \COVER. By \eqref{eq:overlap}, the covariate-overlap penalty equals \(\lambda\) times the average of \(\mathcal C_{ts}\) over all unordered task pairs. This normalization also keeps the effective regularization strength stable as \(T\) changes. When all tasks have the same second-moment matrix, the effective ridge multiplier is \(2\lambda T/(T-1)\), decreasing from \(4\lambda\) at \(T=2\) toward \(2\lambda\).
For the end-to-end risk analysis in Section~\ref{sec:end-to-end-theory},
we optimize \eqref{eq:estimator} over neural-network classes with bounded
outputs and task-specific coefficients with bounded norms. These
restrictions control model complexity; the loss and covariate-overlap
penalty remain unchanged. Appendix~\ref{app:new-task-transfer} shows how
the same overlap construction transfers the fitted source model to a new
task with limited labeled data.

\section{Computation}\label{sec:computation}

This section develops an implementation of \COVER{} suitable for end-to-end training. We introduce an exact auxiliary formulation that avoids explicit matrix inverses, then describe constraint enforcement and the resulting training procedure.

For each pair \(1\le t<s\le T\), introduce an auxiliary coefficient \(a_{ts}\in\mathbb R^d\). By \eqref{eq:overlap}, the estimator in \eqref{eq:estimator} is equivalent to
\begin{revision}
\begin{equation}\label{eq:auxiliary}
\begin{aligned}
&\min_{\substack{g,z,\{\beta_t\}_{t=1}^T\\
\{a_{ts}\}_{1\le t<s\le T}}}\quad
\sum_{t=1}^T\frac1{2Tn_t}
\sum_{i=1}^{n_t}
\{Y_{ti}-f_t(X_{ti})\}^2\\
&\quad+\frac{2\lambda}{T(T-1)}
\sum_{1\le t<s\le T}
\{\|\beta_t-a_{ts}\|_{\widehat\Sigma_t}^2+
\|\beta_s-a_{ts}\|_{\widehat\Sigma_s}^2\},
\qquad \text{subject to }\sum_{t=1}^T\beta_t=0.
\end{aligned}
\end{equation}
\end{revision}
Each taskwise contribution to the pooling cost is evaluated directly from the representation outputs as \(\|\beta_t-a_{ts}\|_{\widehat\Sigma_t}^2=n_t^{-1}\sum_{i=1}^{n_t}\{z(X_{ti})^\top(\beta_t-a_{ts})\}^2\). Training therefore avoids costly matrix operations such as inversion and eigendecomposition. Minimizing \eqref{eq:auxiliary} over \(\{a_{ts}\}_{1\le t<s\le T}\) recovers \eqref{eq:estimator}, so the objectives have the same global minimizers, though gradient descent need not follow the same finite-iteration path.

We optimize \eqref{eq:auxiliary} by gradient-based updates. After each update, we subtract \(T^{-1}\sum_{s=1}^T\beta_s\) from every \(\beta_t\), enforcing the centering constraint exactly. The end-to-end analysis in Section~\ref{sec:end-to-end-theory} additionally restricts the task-specific coefficients to a bounded capacity class; this theoretical constraint can be enforced by a separate norm projection and may be chosen large enough to remain inactive in computation. The fixed-representation analysis omits this norm bound and retains an exact solution under the centering constraint. Algorithm~\ref{alg:cover} summarizes the training procedure.

\begin{algorithm}[t]
\caption{Training \COVER}
\label{alg:cover}
\begin{algorithmic}[1]
\Require \(\{\mathcal D_t\}_{t=1}^T\), \(\lambda\), \(d\)
\State \((g,z,\{\beta_t\}_{t=1}^T,\{a_{ts}\}_{1\le t<s\le T})\leftarrow\operatorname{Initialize}(\lambda,d)\)
\State \(\bar\beta\leftarrow T^{-1}\sum_{s=1}^T\beta_s\), \(\beta_t\leftarrow\beta_t-\bar\beta\), \(t\in[T]\)
\While{not converged}
    \State Sample a task-balanced mini-batch from every task
    \State Take one gradient-based optimization step on \eqref{eq:auxiliary}
    \State \(\bar\beta\leftarrow T^{-1}\sum_{s=1}^T\beta_s\), \(\beta_t\leftarrow\beta_t-\bar\beta\), \(t\in[T]\)
\EndWhile
\State \(\widehat\beta_t\leftarrow\beta_t\), \(t\in[T]\)
\Ensure \(\widehat f_t(x)=g(x)+z(x)^\top\widehat\beta_t\), \(t\in[T]\)
\end{algorithmic}
\end{algorithm}

\begin{revision}
We use task-balanced validation squared error to select the regularization parameter and early-stopping checkpoint. In the simulations, the representation dimension and architecture are prespecified and examined separately by sensitivity analysis; in applications, they may instead be validation selected. We choose \(d\) below the task sample sizes and use normalization or weight decay for numerical stability. With all task pairs included, coefficient storage is \(O(Td+T^2d)\). Beyond the shared neural passes, pairwise operations cost \(O(Td\sum_{t=1}^T n_t)\) per full pass and can be vectorized.
\end{revision}

\section{Theoretical Properties}\label{sec:theory}

We first establish separation and invariance properties, then analyze the coefficient estimator conditional on a fixed \(g\) and \(z\), and finally give an end-to-end excess-risk bound for jointly learned neural functions. This distinction separates exact conditional identities for fixed representations from risk bounds that account for learning the neural functions. Throughout this section, the \(T\) task data sets are mutually independent.

\subsection{Separation and Reparameterization Invariance}\label{sec:identifiability}

\begin{revision}
\begin{proposition}[Pure covariate heterogeneity]\label{prop:pure-covariate}
Fix \(g=g^\ast\) and \(z\). Suppose \(f_t^\ast(x)=g^\ast(x)\) and \(\Sigma_t\succ0\) for every \(t\in[T]\). Then the population criterion over the centered coefficients is uniquely minimized at \(\beta_t^\ast=0\) for all \(t\), regardless of the differences among \(\{\mathbb P_t^X\}_{t=1}^T\). If some \(\Sigma_t\) is singular, every minimizer still satisfies \(z(X)^\top\beta_t^\ast=0\) almost surely under \(\mathbb P_t^X\).
\end{proposition}
\end{revision}

Proposition~\ref{prop:pure-covariate} shows how data integration responds to covariate heterogeneity through the covariate-overlap penalty. If the common component function is misspecified, different covariate distributions may also induce task-specific coefficients that reflect approximation error. The proof is provided in Appendix~\ref{app:theory-proofs}.

\begin{revision}
\begin{proposition}[Reparameterization invariance]\label{prop:invariance}
For any invertible \(L\in\mathbb R^{d\times d}\), define \(z_L(x)=Lz(x)\) and \(\beta_{t,L}=L^{-\top}\beta_t\). Then
\[
z_L(x)^\top\beta_{t,L}=z(x)^\top\beta_t,
\qquad
\Sigma_{t,L}=L\Sigma_tL^\top,
\qquad
\Omega_{ts,L}=L\Omega_{ts}L^\top,
\]
and
\[
\|\beta_{t,L}-\beta_{s,L}\|_{\Omega_{ts,L}}^2
=
\|\beta_t-\beta_s\|_{\Omega_{ts}}^2.
\]
Hence the fitted functions and covariate-overlap penalty values are invariant to invertible changes of representation coordinates, even though the coordinates of \(z\) and \(\beta_t\) themselves are not uniquely determined.
\end{proposition}
\end{revision}

\begin{revision}
This result addresses coordinate nonuniqueness within a fixed shared representation span. Although the coordinates of \(z\) and \(\beta_t\) are not unique, predictions and pairwise data integration remain unchanged under invertible coordinate transformations of that span. The coefficient-norm bound in the end-to-end theory is stated in the chosen coordinates and provides capacity control for the oracle inequalities. The proof is provided in Appendix~\ref{app:theory-proofs}.
\end{revision}

\subsection{Fixed-Representation Bias--Variance Decomposition}\label{sec:fixed-representation-theory}

\begin{revision}
We now condition on a fixed common component function \(g\) and shared representation \(z\). This separates the effect of the covariate-overlap penalty from the approximation and estimation errors associated with learning these two functions.
\end{revision}

\begin{revision}
\begin{assumption}[Fixed-representation coefficient model]\label{ass:coefficient-model}
For every \(t\in[T]\), \(n_t=n\), \(Y_{ti}=g(X_{ti})+z(X_{ti})^\top\beta_t^\ast+\varepsilon_{ti}\), and \(\sum_{t=1}^T\beta_t^\ast=0\). Conditional on all observed covariates, the errors are independent with conditional mean zero and conditional variance \(\sigma^2\).
\end{assumption}
\end{revision}

\begin{revision}
Assumption~\ref{ass:coefficient-model} isolates the coefficient-pooling effect when the fixed functions \(g\) and \(z\) span the task conditional means. Equal sample sizes and homoscedastic errors give the closed-form direction-specific variance below; the estimator and algorithm themselves allow unequal \(n_t\).
\end{revision}

Stack the coefficients as \(\beta=(\beta_1^\top,\ldots,\beta_T^\top)^\top\in\mathbb R^{Td}\), and set \(\widehat D=\operatorname{blockdiag}(\widehat\Sigma_1,\ldots,\widehat\Sigma_T)\). Define the matrix-weighted task Laplacian \(\widehat L\) through
\[
\beta^\top\widehat L\beta
=
\sum_{1\le t<s\le T}\|\beta_t-\beta_s\|_{\widehat\Omega_{ts}}^2.
\]

\begin{revision}
\begin{assumption}[Centered second-moment condition]\label{ass:moment-condition}
Call a stacked vector centered if its task blocks sum to zero. For every nonzero centered \(v\), assume that \(v^\top\widehat Dv>0\). Thus the empirical second-moment matrices identify all coefficient differences considered below.
\end{assumption}
\end{revision}

\begin{revision}
Assumption~\ref{ass:moment-condition} requires rank only on the centered subspace; positive definiteness of every \(\widehat\Sigma_t\) is sufficient but unnecessary. It fails exactly when there exist vectors \(v_t\in\Null(\widehat\Sigma_t)\), not all zero, whose sum is zero. It holds almost surely when \(n\ge d\) and each fixed, nondegenerate continuous task representation has full column rank, but can fail when \(d>n\) or the representation collapses. For fixed \(g\) and \(z\), the condition ensures uniqueness of the centered coefficient minimizer.
\end{revision}

Under Assumption~\ref{ass:moment-condition}, choose a \(\widehat D\)-orthonormal generalized eigenbasis \(v_1,\ldots,v_{(T-1)d}\) of the centered subspace such that
\begin{equation}\label{eq:spectral}
u^\top\widehat Lv_j=\mu_j u^\top\widehat Dv_j
\quad\text{for every centered }u,
\qquad \mu_j\ge0,
\qquad j=1,\ldots,(T-1)d.
\end{equation}
Set \(\kappa=2\lambda/(T-1)\), \(\theta_j=v_j^\top\widehat D\beta^\ast\), and let \(\mathscr X=\{X_{ti}:t\in[T],i\in[n]\}\) collect all observed covariates.

\begin{revision}
\begin{theorem}[Fixed-representation bias--variance]\label{thm:fixed-bias-variance}
Fix \(g\) and \(z\), and let \(\widehat\beta\) minimize \eqref{eq:estimator} over the task-specific coefficients. Under Assumptions~\ref{ass:coefficient-model} and \ref{ass:moment-condition},
\begin{equation}\label{eq:fixed-bias-variance}
\mathbb E\!\left\{\left.\frac1T\sum_{t=1}^T
\|\widehat\beta_t-\beta_t^\ast\|_{\widehat\Sigma_t}^2\,\right|\,\mathscr X\right\}
=\frac1T\sum_{j=1}^{(T-1)d}\left\{
\left(\frac{\kappa\mu_j}{1+\kappa\mu_j}\right)^2\theta_j^2
+\frac{\sigma^2/n}{(1+\kappa\mu_j)^2}\right\}.
\end{equation}
The first term is the direction-specific shrinkage bias, and the second is the direction-specific variance. The design-dependent pair \((\mu_j,\theta_j)\) jointly determines the error: \(\mu_j\) gives the overlap-based pooling strength, while \(\theta_j\) records the true coefficient contrast in the same empirical eigenbasis.
\end{theorem}
\end{revision}

\begin{revision}
Theorem~\ref{thm:fixed-bias-variance} gives a direction-by-direction account of data integration. In direction \(j\), the estimator multiplies the noisy signal coordinate by \((1+\kappa\mu_j)^{-1}\). A larger \(\mu_j\) therefore produces stronger coefficient pooling and a larger variance reduction, but it also increases shrinkage bias when \(\theta_j\ne0\). Directions with \(\mu_j=0\) are not pooled because they lie outside the shared representation subspace encoded by the overlap matrices.
\end{revision}

\begin{revision}
The two extremes clarify the role of the regularization parameter \(\lambda\). At \(\lambda=0\), the covariate-overlap penalty applies no coefficient pooling beyond the centering constraint, and the direction-specific variance is \(\sigma^2/n\). As \(\lambda\) increases, variance decreases monotonically, whereas the squared bias approaches \(\theta_j^2\) for directions with \(\mu_j>0\). Thus, pooling improves prediction when the reduction in variance exceeds the increase in squared bias from shrinking true differences between tasks. The overlap matrices determine how strongly each representation direction is pooled. The proof is provided in Appendix~\ref{app:theory-proofs}.
\end{revision}

The sum of the direction-specific variance multipliers in Theorem~\ref{thm:fixed-bias-variance} can be interpreted as the number of centered coefficient directions that remain effectively unpooled after applying the covariate-overlap penalty.

\begin{corollary}[Shrinkage under equal second moments]\label{cor:equal-moment-shrinkage}
Suppose Assumption~\ref{ass:moment-condition} holds and \(\widehat\Sigma_t=\widehat\Sigma\) for all tasks. Then every generalized eigenvalue equals \(T\), and the denominator in \eqref{eq:fixed-bias-variance} is \(1+2\lambda T/(T-1)\).
\end{corollary}

Under equal empirical second moments, centered coefficient differences share a common shrinkage factor, regardless of representation direction. This clarifies the role of covariate heterogeneity in Theorem~\ref{thm:fixed-bias-variance}: equal second moments give uniform shrinkage, whereas unequal second moments allow pooling strength to vary across directions. The proof is provided in Appendix~\ref{app:theory-proofs}.

\subsection{End-to-End Excess Risk}\label{sec:end-to-end-theory}

This subsection extends the fixed-representation analysis to jointly learned common and task-specific component functions. We first specify the neural function classes and risks, then establish global and localized oracle inequalities.

\subsubsection{Neural Function Classes and Risks}

\begin{revision}
We next allow the common component function and shared representation to be learned from the data. Let \(\mathcal G\) be a class of scalar-valued common component functions and let \(\mathcal Z\) be a class of \(d\)-dimensional shared representations. The two classes serve different statistical roles and need not have the same architecture. They may nevertheless use the same depth and hidden-layer widths. If an implementation shares some lower-layer parameters between \(g\) and \(z\), its admissible function pairs form a subset of \(\mathcal G\times\mathcal Z\), so the result below continues to apply.
\end{revision}

For a fixed coefficient bound \(0<B_\beta<\infty\), define the
capacity-constrained model class
\[
\mathcal H=
\left\{h=(g,z,\beta_1,\ldots,\beta_T):
g\in\mathcal G,\ z\in\mathcal Z,\
\max_{t\in[T]}\|\beta_t\|_2\le B_\beta,\
\sum_{t=1}^T\beta_t=0\right\}.
\]
For \(h\in\mathcal H\), write \(f_{h,t}(x)=g(x)+z(x)^\top\beta_t\). The corresponding predictor class is
\(\mathcal F=\{(f_{h,1},\ldots,f_{h,T}):h\in\mathcal H\}\).

\begin{revision}
\begin{assumption}[Bounded neural classes]\label{ass:neural-classes}
For some \(B_Y,B_g,B_z\in(0,\infty)\), \(|Y|\le B_Y\) almost surely under every task, \(\sup_x|g(x)|\le B_g\) for every \(g\in\mathcal G\), and \(\sup_x\|z(x)\|_2\le B_z\) for every \(z\in\mathcal Z\). The classes \(\mathcal G\) and \(\mathcal Z\) are measurable and have finite Rademacher complexities as defined below.
\end{assumption}
\end{revision}

Assumption~\ref{ass:neural-classes} holds for bounded inputs and responses when fixed-depth networks have bounded layer norms; clipping can enforce the output bounds. It covers norm-constrained ReLU, sigmoid, and tanh networks, while the results require only bounded outputs and finite class complexities.

For the end-to-end analysis, set \(N=Tn\) and
\(B=B_Y+B_g+B_zB_\beta\); then
\(|Y-f_{h,t}(X)|\le B\) uniformly over \(h\in\mathcal H\).
For \(h\in\mathcal H\), define the task-balanced population risk and its empirical counterpart by
\[
\mathcal R(h)=\frac1{2T}\sum_{t=1}^T
\mathbb E_t\{Y-f_{h,t}(X)\}^2,
\qquad
\widehat{\mathcal R}(h)=\frac1{2Tn}\sum_{t=1}^T\sum_{i=1}^n
\{Y_{ti}-f_{h,t}(X_{ti})\}^2.
\]
Let \(\Omega_{ts}(z)\) emphasize that the population overlap matrix is formed from the representation \(z\), and set
\[
\mathcal P_\lambda(h)
=\frac{\lambda}{T(T-1)}
\sum_{1\le t<s\le T}\|\beta_t-\beta_s\|_{\Omega_{ts}(z)}^2.
\]
The empirical penalty \(\widehat{\mathcal P}_\lambda(h)\) is defined analogously using \(\widehat\Omega_{ts}(z)\). For the end-to-end analysis, let
\(\widehat h\in\arg\min_{h\in\mathcal H}
\{\widehat{\mathcal R}(h)+\widehat{\mathcal P}_\lambda(h)\}\).
This is the capacity-constrained version of \eqref{eq:estimator}. The
coefficient constraint can be enforced by projection and may be chosen
large enough to be inactive in a fixed-representation analysis;
accordingly, Theorem~\ref{thm:fixed-bias-variance} retains the exact
unconstrained quadratic solution.

Let \(f^\ast=(f_1^\ast,\ldots,f_T^\ast)\) be the tuple of true conditional means and define
\[
\mathcal E(h)
=\mathcal R(h)-\mathcal R(f^\ast)
=\frac1{2T}\sum_{t=1}^T
\mathbb E_t\{f_{h,t}(X)-f_t^\ast(X)\}^2.
\]
The second equality is the squared-loss projection identity and does not require \(f^\ast\in\mathcal F\): writing \(Y=f_t^\ast(X)+\varepsilon_t\), the remaining cross term vanishes because \(\mathbb E_t(\varepsilon_t\mid X)=0\).

\subsubsection{Uniform Control of the Overlap Matrices}

Let \(\mathscr R_N(\mathcal F)\) denote the task-balanced Rademacher complexity of the predictor class, and let \(\widetilde{\mathscr R}_n(\mathcal Z)\) denote the largest taskwise complexity of the coordinate products \(z_jz_k\). Their precise definitions are collected in Appendix~\ref{app:complexity-notation}. The first quantity controls prediction loss over all \(N=Tn\) observations; the second controls estimation of the task-specific second-moment matrices and does not introduce an additional model class.

For \(0<\delta<1\), define
\begin{equation}\label{eq:covariance-rate}
r_{\Sigma,n}(\delta)
=2d\,\widetilde{\mathscr R}_n(\mathcal Z)
+dB_z^2\sqrt{\frac{2\log(Td^2/\delta)}{n}}.
\end{equation}
The quantity \(r_{\Sigma,n}(\delta)\) controls second-moment estimation errors uniformly over tasks and \(z\in\mathcal Z\), including representations learned from the same observations. Under standard neural-network norm constraints, \(\widetilde{\mathscr R}_n(\mathcal Z)\) is of order \(n^{-1/2}\); Appendix~\ref{app:end-to-end-lemmas} provides the uniform bound and explicit network rates. Although the Moore--Penrose formula in \eqref{eq:overlap} allows singular second-moment matrices, we impose the following uniform full-rank condition to ensure stable estimation of the overlap matrices in the end-to-end analysis.

\begin{revision}
\begin{assumption}[Population second-moment matrices]\label{ass:uniform-spectrum}
For some \(0<m_z\le M_z<\infty\), every \(z\in\mathcal Z\) and \(t\in[T]\) satisfy \(m_z I_d\preceq\Sigma_t(z)\preceq M_z I_d\).
\end{assumption}
\end{revision}

\begin{revision}
Assumption~\ref{ass:uniform-spectrum} is needed only for the uniform nonlinear bound; the structural results allow singular matrices. The upper bound follows from \(\|z(X)\|_2\le B_z\), and the lower bound holds for nondegenerate representations under normalization or a spectral constraint preventing collapse. Thus \(\mathcal Z\) is a normalized, nondegenerate subset of a norm-constrained network class. Under this condition, Lemma~\ref{lem:overlap-stability} in Appendix~\ref{app:end-to-end-lemmas} shows that second-moment accuracy transfers to the overlap matrices, with sensitivity governed by the representation condition number \(M_z/m_z\).
\end{revision}

\subsubsection{Global Oracle Inequality}

\begin{revision}
\begin{theorem}[End-to-end oracle inequality]\label{thm:end-to-end}
Let \(0<\delta<1/2\), and suppose Assumptions~\ref{ass:neural-classes} and \ref{ass:uniform-spectrum} hold. If
\(r_{\Sigma,n}(\delta)\le m_z/2\), then, with probability at least \(1-2\delta\),
\begin{equation}\label{eq:oracle}
\begin{aligned}
\mathcal E(\widehat h)
\le\;&
\underbrace{
\inf_{h\in\mathcal H}
\{\mathcal E(h)+\mathcal P_\lambda(h)\}
}_{\text{approximation and shrinkage}}
+
\underbrace{
8B\,\mathscr R_N(\mathcal F)
}_{\text{neural-class complexity}}\\
&+
\underbrace{
B^2\sqrt{\frac{\log(1/\delta)}{2Tn}}
}_{\text{concentration}}
+
\underbrace{
28\lambda B_\beta^2
\left(\frac{M_z}{m_z}\right)^2
r_{\Sigma,n}(\delta)
}_{\text{overlap-matrix estimation}}.
\end{aligned}
\end{equation}
The labels in \eqref{eq:oracle} separate approximation and regularization bias from the two empirical-process terms and the additional error caused by estimating the overlap matrices.
\end{theorem}
\end{revision}

\begin{revision}
Theorem~\ref{thm:end-to-end} separates the statistical roles of data integration. The regularized oracle term rewards representations and coefficients that fit the conditional means while requiring little pooling distortion. The neural-class and concentration terms are the usual price of learning the predictors. The final term is specific to \COVER: it is the cost of estimating, from the same data, the overlap matrices that determine the covariate-overlap penalty.
\end{revision}

\begin{revision}
The bound also clarifies what the end-to-end result does and does not establish. Setting \(\lambda=0\) removes both the population covariate-overlap penalty and the overlap-estimation term, yielding the corresponding unregularized multi-task benchmark over the same neural class. For \(\lambda>0\), the bound is favorable when the task differences are small along the shared representation subspaces and \(r_{\Sigma,n}(\delta)\) is controlled. Theorem~\ref{thm:fixed-bias-variance}, rather than the global complexity term here, gives the explicit variance reduction produced by data integration. The proof of Theorem~\ref{thm:end-to-end} is provided in Appendix~\ref{app:theory-proofs}.
\end{revision}

The global bound requires no further noise assumption but does not accelerate when the best regularized predictor has small risk. A localized refinement gives this improvement and makes the dependence on \(T\), \(n\), \(d\), and the neural classes explicit.

\subsubsection{Optimistic Oracle Inequality}

Let \(\mathscr G_N(\mathcal G)\) and \(\mathscr G_N(\mathcal Z)\) be the Gaussian complexities of \(\mathcal G\) and \(\mathcal Z\). Appendix~\ref{app:complexity-notation} gives their worst-case definitions. Define the localized prediction scale \(r_N\) by
\begin{equation}\label{eq:local-scale}
\sqrt{r_N}=
\left\{
\mathscr G_N(\mathcal G)
+B_\beta\mathscr G_N(\mathcal Z)
+B_zB_\beta\sqrt{\frac dn}
\right\}\log^2(eN)
+\frac{(B_g+B_zB_\beta)\log(eN)+B}{\sqrt N}.
\end{equation}
The common component and shared representation in \eqref{eq:local-scale} use all \(N=Tn\) observations, whereas the coefficient term remains at the within-task scale \(n\). Proposition~\ref{prop:local-complexity} in Appendix~\ref{app:local-analysis} shows that \(r_N\) bounds the localized squared-loss fixed point; its proof in Appendix~\ref{app:supplementary-proofs} gives the local Rademacher and Gaussian-comparison details.

Let \(h_\lambda^\ast\in\arg\min_{h\in\mathcal H}
\{\mathcal R(h)+\mathcal P_\lambda(h)\}\) be a population-penalized
oracle, and let
\(\Delta_{\lambda,n}(\delta)=14\lambda B_\beta^2
(M_z/m_z)^2r_{\Sigma,n}(\delta)\) denote the error caused by estimating
the overlap matrices.

\begin{revision}
\begin{theorem}[Optimistic end-to-end oracle inequality]\label{thm:optimistic-oracle}
Let \(0<\delta<1/3\), and suppose Assumptions~\ref{ass:neural-classes} and \ref{ass:uniform-spectrum} hold. If \(r_{\Sigma,n}(\delta)\le m_z/2\), then, with probability at least \(1-3\delta\),
\begin{equation}\label{eq:optimistic-oracle}
\begin{aligned}
\mathcal E(\widehat h)
\le\;&
\underbrace{
\mathcal E(h_\lambda^\ast)+\mathcal P_\lambda(h_\lambda^\ast)
}_{\text{approximation and shrinkage}}
+\underbrace{
C\Delta_{\lambda,n}(\delta)
}_{\text{overlap-matrix estimation}}\\
&+
\underbrace{
C\left[
\sqrt{
\{\mathcal R(h_\lambda^\ast)+\mathcal P_\lambda(h_\lambda^\ast)
+\Delta_{\lambda,n}(\delta)\}
\left\{r_N+\frac{B^2\log(1/\delta)}N\right\}
}
+r_N+\frac{B^2\log(1/\delta)}N
\right]
}_{\text{localized estimation}} .
\end{aligned}
\end{equation}
where \(C>0\) is a universal constant.
\end{theorem}
\end{revision}

\begin{revision}
Theorem~\ref{thm:optimistic-oracle} refines Theorem~\ref{thm:end-to-end} in low-risk regimes. When the raw oracle risk is not small, the leading localized term has the usual square-root behavior. When the raw oracle risk and overlap-estimation error are both comparable to \(r_N+B^2\log(1/\delta)/N\), the stochastic prediction term has this faster order. The covariate-overlap penalty affects the bound in two distinct ways: it changes the population oracle through data integration, and estimation of its overlap matrices contributes \(\Delta_{\lambda,n}(\delta)\). The fixed-representation result in Theorem~\ref{thm:fixed-bias-variance} further shows how the same data integration reduces variance direction by direction. Appendix~\ref{app:theory-proofs} gives the proof. Both inequalities concern an exact global minimizer. If numerical optimization is within \(\varepsilon_{\mathrm{opt}}\) of the empirical infimum, \eqref{eq:oracle} gains \(\varepsilon_{\mathrm{opt}}\); in \eqref{eq:optimistic-oracle}, the proof adds \(C\varepsilon_{\mathrm{opt}}\) and includes \(\varepsilon_{\mathrm{opt}}\) inside the localized square root.
\end{revision}

For norm-constrained ReLU networks, Corollary~\ref{cor:explicit-neural-rate} gives an explicit rate separating the combined-sample scale \(Tn\) for shared functions from the within-task scale \(n\) for task-specific coefficients and overlap matrices. Network-norm and coordinatewise complexity details are in Appendix~\ref{app:local-analysis}.

\section{Experiments}\label{sec:experiments}

We test prediction under joint heterogeneity, the overlap mechanism in Section~\ref{sec:overlap}, and scaling with the number of tasks and covariates, followed by a Genotype--Tissue Expression (GTEx) central-nervous-system analysis. Tuning uses validation data only, test data are reserved for final evaluation, and simulations use 100 independent repetitions unless stated otherwise. Code and implementation details for all reported experiments, including tuning and optimization settings, are available at {\urlstyle{same}\url{https://github.com/YS-stat/COVER-MTL}}.

\subsection{Experimental Setup}\label{sec:experimental-setup}

The main simulation has \(T=48\) tasks arranged in six covariate groups, with eight tasks per group. Each task has 100 training, 200 validation, and 3,000 test observations. We set the input and representation dimensions to \(p=d=24\). Let \(U=Q^\top X\), where \(Q\) is a random orthogonal matrix shared by all tasks. Each covariate group has variance one on four group-specific coordinates of \(U\) and variance 0.001 on the remaining coordinates; the six groups partition the 24 coordinates. The true shared representation is \(z^\ast(X)=\tanh(U)\), applied coordinatewise, and the common component function is
\[
g^\ast(X)=0.8\sin(U_1)+0.5(U_2)_+-0.4(-U_3)_++0.3U_4U_5.
\]
Responses are generated from \(Y=g^\ast(X)+z^\ast(X)^\top\beta_t^\ast+\varepsilon\), where \(\varepsilon\sim N(0,1)\). Under joint heterogeneity, tasks in the same covariate group share a coefficient prototype in that group's high-variance representation subspace and receive centered within-group perturbations. The between-group component scale is one, while the within-group component scale is 0.20 in the moderate setting and 0.30 in the strong setting. This construction makes the two forms of heterogeneity distinct: the second-moment matrices determine the represented directions, whereas \(\beta_t^\ast\) determines how the conditional mean changes.

Three controls isolate the two forms of heterogeneity. The homogeneous setting has equal second moments and \(\beta_t^\ast=0\). Covariate Only retains the six covariate groups but sets \(\beta_t^\ast=0\). Posterior Only uses identity second moments and independent centered coefficients rescaled to unit task-balanced prediction root mean squared magnitude. These settings test null behavior under pure covariate heterogeneity and performance when second moments are equal.

\begin{revision}
We compare seven methods. Pooled learning (Pool) fits one common ReLU regression network. Single-task learning (STL) fits an independent network to each task. Hard parameter sharing (HPS) uses the decomposition in \eqref{eq:model} without the covariate-overlap penalty and is the direct neural analogue of classical shared-parameter MTL \citep{caruana1997multitask}. Multi-gate mixture-of-experts (MMoE) uses four shared experts with task-specific gates and towers \citep{ma2018modeling}. Adaptive and robust multi-task learning (ARMUL) applies the group-correction penalty of \citet{duan2023adaptive} to the learned shared representation from HPS. Fused lasso approach in regression coefficients clustering (FLARCC) applies the coefficient-fusion component of \citet{tang2016fused} to the same learned shared representation; its variable-selection term is omitted so the comparison isolates cross-task fusion. Average-Moment, used only in the mechanism study, has the same model and optimization as \COVER{} but replaces \(\Omega_{ts}\) by \((\Sigma_t+\Sigma_s)/2\).
\end{revision}

\begin{revision}
To make the comparison reflect the data-integration rule rather than differences in data use or architecture search, we keep the common ingredients fixed whenever the method permits. All methods use the same training, validation, and test splits and the same task-balanced validation criterion. HPS and \COVER{} use exactly the same \(g\) and \(z\) networks, representation dimension, and initial HPS fit; ARMUL and FLARCC operate on that same learned shared representation and refit the common component function after estimating their task-specific coefficients. Pool and STL use the same common-component network architecture as HPS, whereas MMoE retains its method-specific expert and gating architecture. All neural components are trained with the same optimizer settings and early-stopping rule. Regularization parameters for \COVER{}, ARMUL, and FLARCC are selected separately by validation loss over prespecified grids, so no method is assigned a test-informed value.
\end{revision}

\begin{revision}
The frozen simulation setup uses one hidden ReLU layer of width 32 for \(g\), one hidden ReLU layer of width 48 and output dimension \(d=24\) for \(z\), and four experts for MMoE. Optimization uses AdamW with learning rate \(10^{-3}\), gradient clipping at 5, and task-balanced mini-batches of 64 observations per task. Numerical capacity is controlled by weight decay and early stopping, while the bounded classes in Theorems~\ref{thm:end-to-end} and~\ref{thm:optimistic-oracle} provide conditions for the finite-sample analysis. The initial neural fit uses at most 2,500 updates, and each regularized path uses at most 1,000 additional updates; the stopping checkpoint in both stages is selected by validation loss. Each positive-\(\lambda\) \COVER{} path starts from the selected HPS fit, with \(a_{ts}\) initialized at \((\beta_t+\beta_s)/2\). The \COVER{} regularization parameter is selected from \(\{0,0.01,0.03,0.1,0.3,1,3,10,30,100\}\). We prespecify the architecture and \(d\) across the 100 repetitions rather than selecting them after observing test performance. Appendix~\ref{sec:additional-sensitivity} separately varies \(d\), network width, depth, and initialization and shows that the comparison is stable across these choices. This sensitivity analysis supports the reported base configuration while keeping the main comparison transparent and computationally reproducible.
\end{revision}

The primary metric is task-balanced excess mean squared error,
\[
\frac1T\sum_{t=1}^T\frac1{n_{\mathrm{te}}}
\sum_{i=1}^{n_{\mathrm{te}}}
\{\widehat f_t(X_{ti}^{\mathrm{te}})-f_t^\ast(X_{ti}^{\mathrm{te}})\}^2.
\]
We also record worst-task and task-specific component errors, selected regularization parameters, computation, and learned overlap. Tables report Monte Carlo standard deviations and figures report 95\% confidence intervals.

\subsection{Performance across Heterogeneity Regimes}\label{sec:heterogeneity-results}

\begin{table}[t]
\centering
\footnotesize
\setlength{\tabcolsep}{3.2pt}
\begin{tabular}{lccccc}
\toprule
Method & Homogeneous & Covariate only & Posterior only & Joint--moderate & Joint--strong \\
\midrule
Pool & \underline{0.1022 (0.0064)} & \textbf{0.0187 (0.0022)} & 1.1335 (0.0110) & 0.1065 (0.0055) & 0.1581 (0.0055) \\
STL & 0.3912 (0.0088) & 0.0555 (0.0051) & \textbf{0.6374 (0.0127)} & 0.1537 (0.0120) & 0.1575 (0.0120) \\
HPS & 0.2280 (0.0136) & 0.0333 (0.0041) & 0.7708 (0.0246) & 0.1167 (0.0085) & 0.1298 (0.0086) \\
MMoE & 0.2651 (0.0189) & 0.0429 (0.0074) & 0.7685 (0.0203) & 0.1324 (0.0112) & 0.1428 (0.0109) \\
ARMUL & 0.1042 (0.0061) & 0.0212 (0.0030) & 0.6878 (0.0235) & \underline{0.1005 (0.0066)} & \underline{0.1129 (0.0075)} \\
FLARCC & 0.1043 (0.0063) & 0.0208 (0.0028) & 0.7331 (0.0228) & 0.1190 (0.0081) & 0.1357 (0.0099) \\
COVER & \textbf{0.1022 (0.0061)} & \underline{0.0193 (0.0023)} & \underline{0.6573 (0.0185)} & \textbf{0.0885 (0.0057)} & \textbf{0.1058 (0.0057)} \\
\bottomrule
\end{tabular}
\caption{Task-balanced excess mean squared error across heterogeneity regimes. Entries are means (standard deviations) over 100 repetitions. The best and second-best results in each column are shown in bold and underlined, respectively.}
\label{tab:simulation-main}
\end{table}

Table~\ref{tab:simulation-main} separates the roles of the two forms of heterogeneity. In the Homogeneous setting, \COVER{} and Pool are essentially tied, showing that validation can suppress unnecessary task-specific component functions. In Covariate Only, Pool is best and \COVER{} is second, with only a small difference. This is the desired null behavior: unequal covariate distributions do not by themselves justify different conditional means. In Posterior Only, STL is best and \COVER{} is second. Because the task-specific coefficients in this control are independent and all tasks have the same second-moment matrix, covariate-adaptive data integration has little structural information to exploit.

The main benefit appears when both forms of heterogeneity are present. Under Moderate Joint Heterogeneity, \COVER{} reduces excess MSE by 12.0\% relative to ARMUL, the second-best method, and by 24.2\% relative to HPS. Under Strong Joint Heterogeneity, the corresponding reductions are 6.3\% and 18.5\%. Pool deteriorates as conditional-mean heterogeneity increases, whereas STL gives up the variance reduction from data integration. ARMUL improves on HPS through global shrinkage, but its penalty does not adapt the direction of shrinkage to the pairwise shared representation subspaces. The results therefore agree with the intended division of labor: conditional-mean heterogeneity determines the differences that must be estimated, while covariate heterogeneity determines which differences can be reliably pooled. Appendix~\ref{sec:additional-secondary} reports worst-task and task-specific component errors, and Appendix~\ref{sec:additional-random} removes the designed alignment between the two forms of heterogeneity.

\subsection{How Covariate Overlap Controls Integration}\label{sec:mechanism-results}

We compare HPS, Average-Moment, and \COVER{} along two one-dimensional paths. One fixes the task-specific components and changes the weak variance from 0.001 to 1, moving normalized overlap from about 0.17 to 1. The other fixes the low-overlap second moments and varies the between-group component scale from 0 to 1. Its within-group scale remains 0.30, so zero removes only the between-group component.

\begin{figure}[t]
\centering
\includegraphics[width=0.98\textwidth]{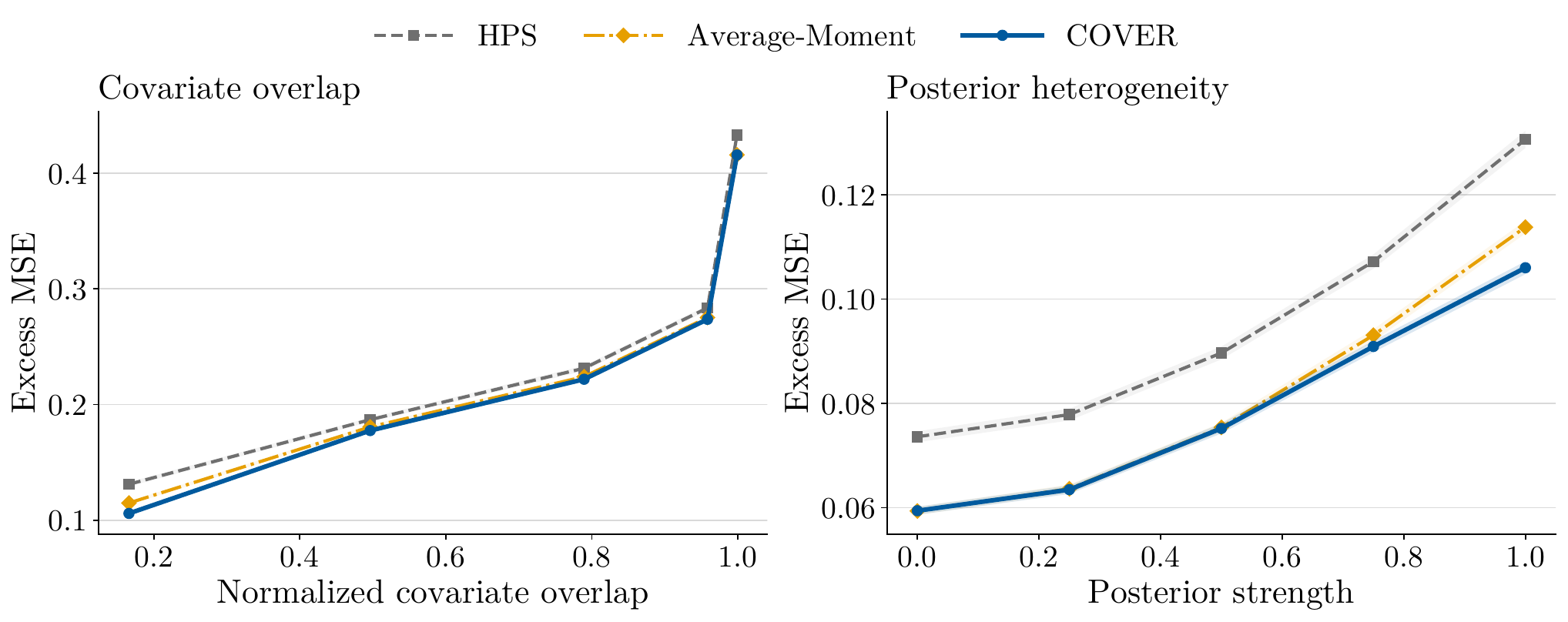}
\caption{Mechanism experiment. The left panel varies covariate overlap while holding the task-specific component structure fixed; the right panel varies the between-group component strength while holding the second-moment structure fixed. Curves are Monte Carlo means with 95\% confidence intervals over 100 repetitions.}
\label{fig:mechanism-main}
\end{figure}

Figure~\ref{fig:mechanism-main} shows that the distinction between the two penalties matters most when overlap is limited. At the smallest overlap, \COVER{} has lower excess MSE than both HPS and Average-Moment. As the second-moment matrices become equal, \(\Omega_{ts}\) approaches the common prediction metric and \COVER{} converges to Average-Moment, as predicted by Corollary~\ref{cor:penalty-comparison}. Along the conditional-mean axis, the gain over HPS becomes more visible as between-group differences increase. The absolute error along the covariate axis also changes with the task-specific prediction norms; comparisons should therefore be made among methods at a fixed horizontal coordinate, rather than interpreting the vertical trend as a causal effect of overlap. Appendix~\ref{sec:additional-mechanism} reports the corresponding normalized-overlap and task-specific component diagnostics.

\subsection{Scalability and Computation}\label{sec:scaling-results}

We vary \(T\in\{24,96,192\}\) at \(p=24\) and \(p\in\{50,100,200\}\) at \(T=48\). Task scaling retains six covariate groups and increases their sizes; dimension scaling rotates the 24-dimensional predictive block and appends independent noise coordinates. Each task retains 100 training and 200 validation observations. All settings use Strong Joint Heterogeneity and repeat the full validation and fitting workflow.

\begin{figure}[t]
\centering
\includegraphics[width=0.99\textwidth]{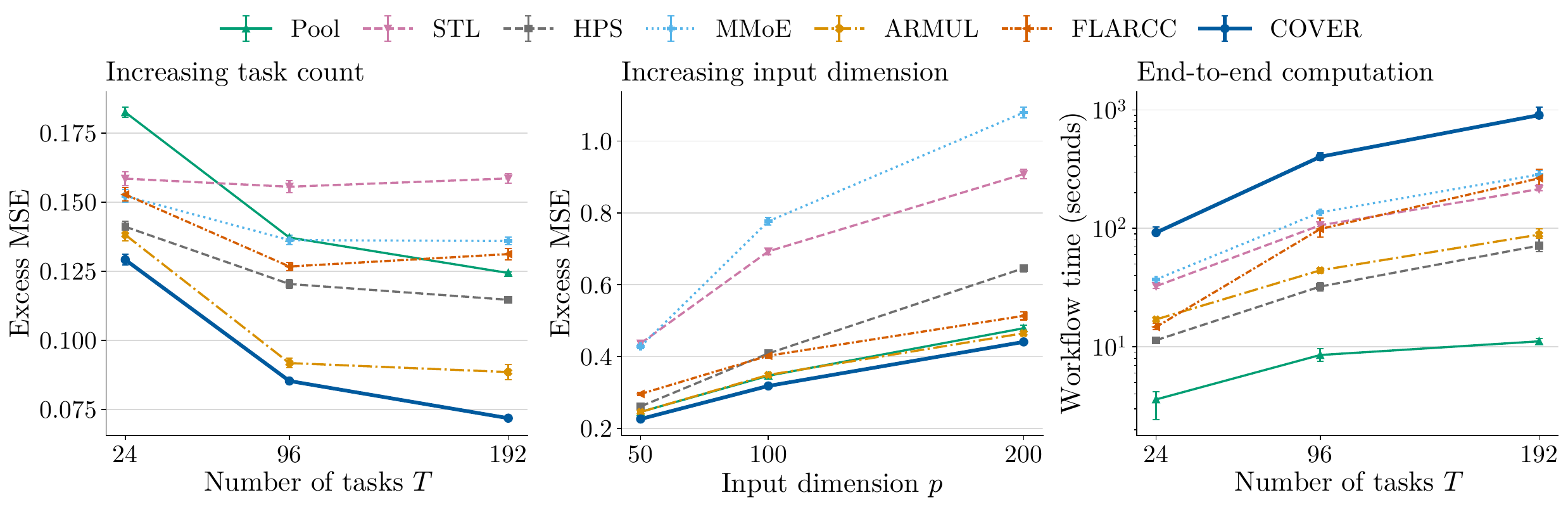}
\caption{Scaling with the number of tasks and input dimension. The first two panels show task-balanced excess mean squared error with 95\% confidence intervals. The last panel shows median end-to-end workflow time with interquartile ranges as the number of tasks increases; the vertical axis is logarithmic.}
\label{fig:scaling-main}
\end{figure}

\begin{revision}
As shown in Figure~\ref{fig:scaling-main}, \COVER{} has the lowest excess MSE at every displayed task count and input dimension. At \(T=192\), its error is 0.0718, compared with 0.0885 for ARMUL and 0.1147 for HPS. At \(p=200\), its error is 0.4409, compared with 0.4650 for ARMUL and 0.4789 for Pool. The improvement with more tasks is consistent with pooling more observations within each repeatedly occurring covariate group. The full \COVER{} path is computationally more expensive than HPS because it evaluates several regularization-parameter values and maintains pairwise auxiliary coefficients. The observed workflow remains feasible through 192 tasks, but the dense pairwise implementation is not intended to be linear in \(T\). Appendix~\ref{sec:additional-resources} reports memory, parameter counts, and selected regularization parameters at the largest settings.
\end{revision}

\raggedbottom
\subsection{GTEx Central-Nervous-System Analysis}\label{sec:gtex}

We analyze version 8 of the GTEx project \citep{gtex2020atlas}. The tasks are 11 central-nervous-system tissues, including cervical spinal cord, and the two responses are the expression levels of \textit{JAM2} and \textit{SH2D2A}. \change{We chose this example exploratorily as a focused central-nervous-system gene-regulation benchmark: the two responses belong to coexpression set MODULE 137, whose remaining genes form the candidate predictor set.} The data contain 2,306 tissue samples from 381 donors, with 152--255 samples per tissue. We transform expression as \(\log_2(\mathrm{TPM}+1)\); the resulting analysis matrix has no missing entries.

\begin{revision}
Within each training fold, we retain the 500 genes with largest task-balanced variance, standardize them by task-balanced training moments, and whiten the leading 30 pooled principal components; responses are standardized within tissue. Donors are never split across folds, although observations across tissue tasks may remain dependent because many donors contribute to several tissues. For each test fold, the next fold is validation and the other three are training. We repeat donor-level five-fold cross-validation 20 times with independent fold and model seeds, reporting standard deviations across 20 complete out-of-fold estimates. The common-component and shared-representation networks each use one ReLU hidden layer of width 32, with \(d=8\). Validation selects the regularization parameter from zero and eight logarithmically spaced values between 0.01 and 30. HPS and every \COVER{} candidate use identical initializations, training seeds, and optimization budgets. MMoE uses three experts. ARMUL and FLARCC use the learned shared representation from HPS in the corresponding fold and their own validation-selected penalties. Performance is measured by tissue-balanced standardized prediction mean squared error. The TissueMean baseline predicts each tissue's training mean, whereas GlobalMean uses the equally weighted average of the 11 tissue-specific training means.
\end{revision}

\begin{table}[t]
\centering
\small
\begin{tabular}{lccc}
\toprule
Method & JAM2 & SH2D2A & Overall \\
\midrule
Global Mean & 1.2980 (0.0098) & 1.6491 (0.0614) & 1.4736 (0.0323) \\
Tissue Mean & 1.0430 (0.0108) & 1.2373 (0.0641) & 1.1401 (0.0331) \\
\midrule
Pool & 0.3175 (0.0092) & \textbf{0.6786 (0.0400)} & 0.4981 (0.0212) \\
STL & 0.3220 (0.0098) & 0.7845 (0.0524) & 0.5532 (0.0275) \\
HPS & 0.2845 (0.0099) & 0.6918 (0.0440) & 0.4882 (0.0207) \\
MMoE & 0.3139 (0.0170) & 0.7503 (0.0565) & 0.5321 (0.0289) \\
ARMUL & \textbf{0.2806 (0.0088)} & 0.6831 (0.0435) & \underline{0.4819 (0.0198)} \\
FLARCC & 0.2858 (0.0104) & 0.6887 (0.0447) & 0.4872 (0.0211) \\
COVER & \underline{0.2808 (0.0099)} & \underline{0.6802 (0.0429)} & \textbf{0.4805 (0.0203)} \\
\bottomrule
\end{tabular}
\caption{Prediction performance in the GTEx central-nervous-system analysis. Entries are standardized mean squared errors, reported as means (standard deviations) over 20 repeated five-fold cross-validation partitions.}
\label{tab:gtex-main}
\end{table}

\begin{revision}
All learned methods substantially improve on the mean baselines in Table~\ref{tab:gtex-main}. \COVER{} has the best response-averaged performance and remains among the leaders for both responses. ARMUL and \COVER{} perform best for \textit{JAM2}; Pool is best and \COVER{} second for \textit{SH2D2A}, consistent with weaker cross-tissue posterior heterogeneity. Appendix~\ref{sec:additional-gtex} reports tissue-specific results: \COVER{} ranks first or second in 12 of 22 response--tissue combinations and among the top three in 19.
\end{revision}

\begin{figure}[!tbp]
\centering
\includegraphics[width=0.94\textwidth]{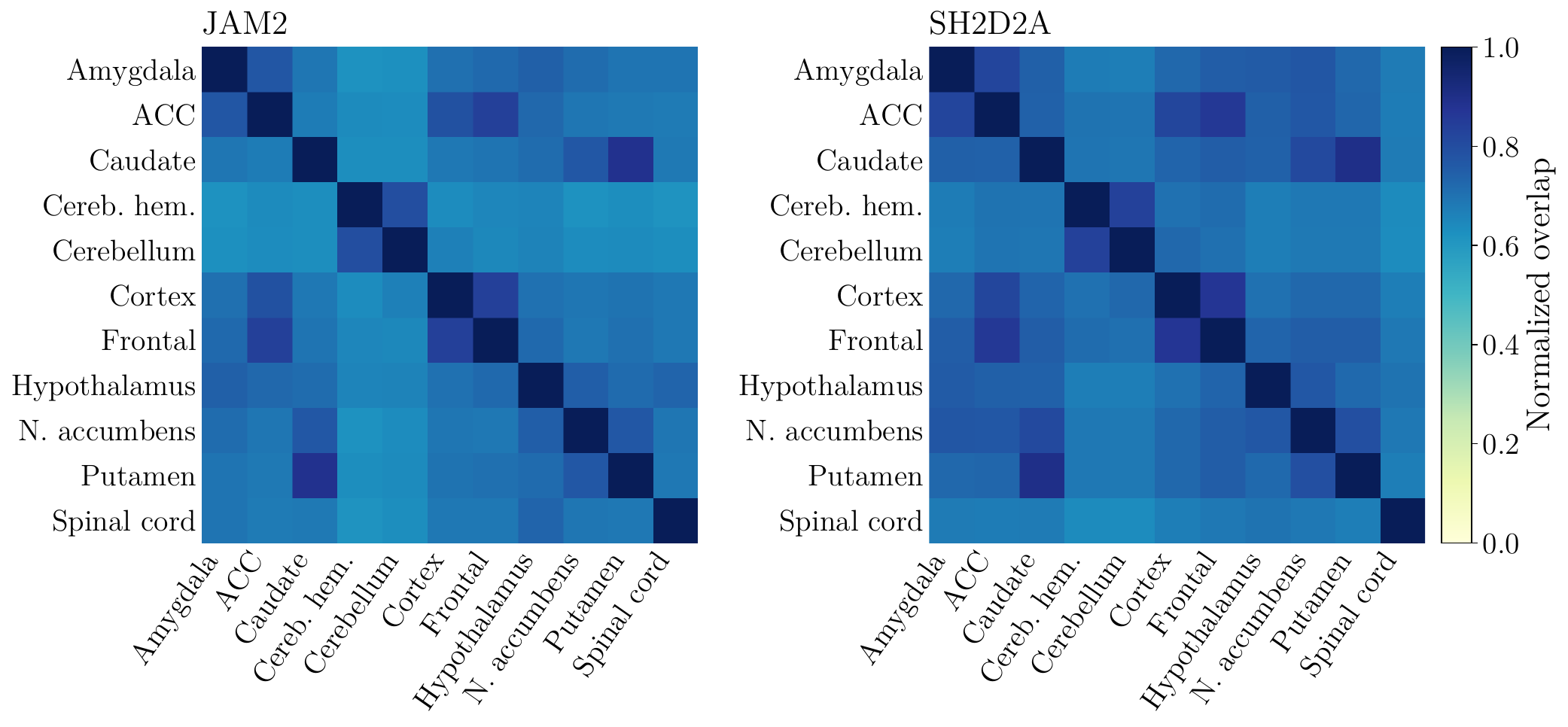}
\caption{Pairwise normalized overlap in the GTEx analysis, conditional on the shared representation fitted by \COVER{} and averaged over folds and repetitions. Each entry is the mean generalized eigenvalue of the overlap matrix relative to the corresponding pairwise average second-moment matrix; larger values indicate stronger relative covariate overlap.}
\label{fig:gtex-overlap}
\end{figure}

Figure~\ref{fig:gtex-overlap} shows different relative overlap strengths across tissue pairs. Anatomically related pairs, including caudate--putamen and cortex--frontal cortex, have relatively strong overlap. This pattern agrees qualitatively with the homogeneous neocortical patterns and similarity of nearby cortical regions reported by \citet{hawrylycz2012atlas}, and with the strong caudate--putamen association in GTEx aging responses found by \citet{brinkmeyer2016aging}.

\section{Discussion}\label{sec:discussion}

We developed \COVER{}, a multi-task learning framework that uses taskwise covariate overlap to determine how task-specific coefficients are integrated. The method separates the roles of the two forms of heterogeneity: covariate heterogeneity determines where integration is supported, whereas posterior heterogeneity determines the coefficient differences being regularized. Its auxiliary formulation permits end-to-end training without matrix inversion, and the theoretical and empirical results characterize the resulting balance between shrinkage bias and variance reduction.

\begin{revision}
Several directions merit further study. First, \COVER{} provides its strongest protection against over-pooling when an outlying task also has limited covariate overlap. For high-overlap posterior outliers, the overlap matrix could be combined with robust fusion \citep{duan2023adaptive} or a learned sparse task graph to accommodate partially shared task structure while retaining direction-specific integration. Second, the dense pairwise formulation requires storage quadratic in the number of tasks. Overlap-guided graph sparsification or stochastic task-pair sampling may extend \COVER{} to substantially larger task collections; an important question is when these approximations preserve the statistical benefit of the full criterion. Finally, the present development targets conditional means under squared loss. Extensions to generalized responses, quantiles, survival outcomes, or distributional prediction would require replacing the second-moment geometry with a suitable loss-dependent predictive geometry. Developing uncertainty quantification for this learned geometry, especially for new-task transfer, is another useful direction.
\end{revision}

\flushbottom
\appendix

\vspace{0.5\baselineskip}
\noindent\emph{Appendix organization.}
Appendix~\ref{app:additional-experiments} gives additional experiments;
Appendix~\ref{app:multiway-pooling} extends the criterion to multiway pooling;
Appendix~\ref{app:supp-theory} develops supplementary theory;
Appendix~\ref{app:new-task-transfer} studies a new but related task; and
Appendix~\ref{app:main-proofs} proves the main-text results.

\raggedbottom
\section{\change{Additional Experiments}}\label{app:additional-experiments}

This appendix supplements the primary comparisons. Protocols, splits, and validation rules match Section~\ref{sec:experiments} unless stated otherwise.

\subsection{\change{Secondary Recovery Metrics}}\label{sec:additional-secondary}

This experiment uses the five DGPs, sample sizes, methods, and validation rules of Section~\ref{sec:heterogeneity-results}. For a test set of size \(n_{\mathrm{te}}\), worst-task excess MSE is the largest task-specific error. To assess posterior-heterogeneity recovery, define \(q_t^\ast=f_t^\ast-g^\ast\) and \(\widehat q_t=\widehat f_t-T^{-1}\sum_{s=1}^T\widehat f_s\). The diagnostics are
\[
\max_{1\le t\le T}\frac1{n_{\mathrm{te}}}\sum_{i=1}^{n_{\mathrm{te}}}
\{\widehat f_t(X_{ti}^{\mathrm{te}})-f_t^\ast(X_{ti}^{\mathrm{te}})\}^2,
\qquad
\frac1T\sum_{t=1}^T\frac1{n_{\mathrm{te}}}\sum_{i=1}^{n_{\mathrm{te}}}
\{\widehat q_t(X_{ti}^{\mathrm{te}})-q_t^\ast(X_{ti}^{\mathrm{te}})\}^2.
\]

\begin{table}[!tbp]
\centering
\scriptsize
\setlength{\tabcolsep}{3.5pt}
\begin{tabular}{lccccc}
\toprule
Method & Homogeneous & Covariate only & Posterior only & Joint--moderate & Joint--strong \\
\midrule
\multicolumn{6}{l}{\textit{Panel A: Worst-task excess MSE}} \\
Pool & 0.1134 (0.0074) & 0.0514 (0.0080) & 1.9356 (0.1818) & 0.2182 (0.0335) & 0.3781 (0.0623) \\
STL & 0.5272 (0.0407) & 0.1824 (0.0401) & 0.9960 (0.1007) & 0.3490 (0.0750) & 0.3662 (0.0855) \\
HPS & 0.3295 (0.0473) & 0.1095 (0.0285) & 1.2673 (0.1249) & 0.2549 (0.0510) & 0.2868 (0.0519) \\
MMoE & 0.4376 (0.0745) & 0.1643 (0.0492) & 1.2123 (0.1208) & 0.2852 (0.0553) & 0.3130 (0.0561) \\
ARMUL & 0.1156 (0.0069) & 0.0635 (0.0183) & 1.1335 (0.1040) & 0.2096 (0.0391) & 0.2368 (0.0415) \\
FLARCC & 0.1160 (0.0069) & 0.0586 (0.0125) & 1.1723 (0.1098) & 0.2421 (0.0440) & 0.2836 (0.0464) \\
COVER & 0.1134 (0.0068) & 0.0536 (0.0085) & 1.0600 (0.0986) & 0.1697 (0.0273) & 0.2164 (0.0317) \\
\midrule
\multicolumn{6}{l}{\textit{Panel B: Task-specific component MSE}} \\
Pool & 0.0000 (0.0000) & 0.0000 (0.0000) & 1.0002 (0.0038) & 1.0393 (0.0061) & 1.0892 (0.0063) \\
STL & 0.1415 (0.0103) & 0.0443 (0.0066) & 0.4598 (0.0133) & 0.2296 (0.0187) & 0.2328 (0.0190) \\
HPS & 0.0445 (0.0148) & 0.0327 (0.0094) & 0.6462 (0.0222) & 0.2738 (0.0388) & 0.2852 (0.0391) \\
MMoE & 0.1110 (0.0164) & 0.0610 (0.0171) & 0.6147 (0.0189) & 0.2102 (0.0254) & 0.2202 (0.0265) \\
ARMUL & 0.0000 (0.0000) & 0.0030 (0.0040) & 0.5668 (0.0211) & 0.3913 (0.0788) & 0.3762 (0.0725) \\
FLARCC & 0.0000 (0.0003) & 0.0017 (0.0028) & 0.6116 (0.0196) & 0.3479 (0.0578) & 0.3541 (0.0572) \\
COVER & 0.0004 (0.0003) & 0.0023 (0.0012) & 0.5327 (0.0158) & 0.7048 (0.0383) & 0.6860 (0.0761) \\
\bottomrule
\end{tabular}
\caption{Secondary recovery metrics across heterogeneity regimes. Entries are means (standard deviations) over 100 repetitions.}
\label{tab:simulation-secondary}
\end{table}

Panel A of Table~\ref{tab:simulation-secondary} shows that the average improvement of \COVER{} is not obtained by sacrificing the most difficult task: it has the smallest worst-task excess error in both joint-heterogeneity settings. Panel B evaluates the simulator-specific decomposition into common and task-specific component functions. Unlike individual coordinates of \(z\) and \(\beta_t\), \(\widehat q_t\) is defined entirely through fitted functions and is invariant to invertible representation reparameterizations. It is nevertheless a more stringent target than \(\widehat f_t\): errors in the estimated common and task-specific component functions can partially cancel in their sum. Component-recovery error and prediction error therefore need not rank methods identically.

\subsection{\change{Mechanism Diagnostics}}\label{sec:additional-mechanism}

Following the joint-heterogeneity DGP, the covariate axis fixes task-specific components while varying the weak variance over \(0.001,0.03,0.1,0.3,1\). The posterior axis fixes this variance at \(0.001\) and the within-group scale at \(0.30\), while varying the between-group scale over \(0,0.25,0.5,0.75,1\).

We report two population diagnostics. Let \(\overline\Sigma=T^{-1}\sum_{t=1}^T\Sigma_t\), let \(r=\operatorname{rank}(\overline\Sigma)\), and write \(\Delta_{ts}=\beta_t^\ast-\beta_s^\ast\). The normalized overlap and supported-contrast fraction are
\[
\frac{2}{T(T-1)}\sum_{t<s}\frac{\operatorname{tr}(\overline\Sigma^\dagger\Omega_{ts})}{r},
\qquad
\frac{\sum_{t<s}\|\Delta_{ts}\|_{\Omega_{ts}}^2}
{\sum_{t<s}\|\Delta_{ts}\|_{(\Sigma_t+\Sigma_s)/2}^2},
\]
respectively. The first is the normalized overlap relative to the pooled second-moment matrix; it increases to one as the task-specific second-moment matrices become equal. The second measures the share of task-specific component contrast energy lying in directions supported by both tasks. It is small when large task differences occur mainly in weakly shared directions.

\begin{figure}[!tbp]
\centering
\includegraphics[width=0.97\textwidth]{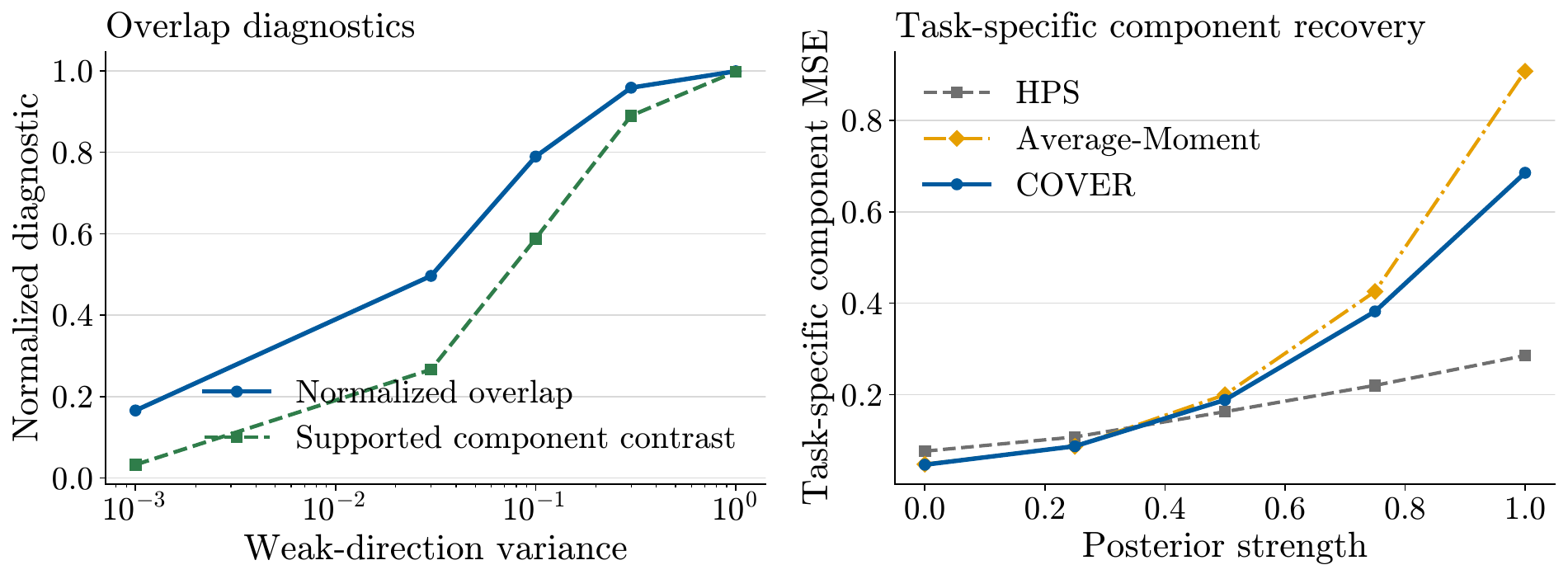}
\caption{Diagnostics for the mechanism experiment. The left panel tracks normalized overlap relative to the pooled second-moment matrix and the fraction of task-specific component contrast energy supported by the overlap matrices. The right panel reports task-specific component recovery as posterior heterogeneity increases. Curves are Monte Carlo means with 95\% confidence intervals.}
\label{fig:mechanism-diagnostics}
\end{figure}

The left panel of Figure~\ref{fig:mechanism-diagnostics} confirms that normalized overlap and the supported fraction increase monotonically along the covariate axis. The right panel shows that prediction and component recovery need not rank methods alike because component errors can offset in the final predictor. Together with Figure~\ref{fig:mechanism-main}, these diagnostics connect the gains to the penalty rather than an architectural change.

\subsection{\change{Random Alignment of the Two Forms of Heterogeneity}}\label{sec:additional-random}

This experiment follows the moderate joint-heterogeneity DGP with \(T=48\), \(p=d=24\), six covariate groups, and 100, 200, and 3,000 training, validation, and test observations per task. Instead of placing coefficient prototypes in group-supported subspaces, it draws coefficients independently of covariate groups, centers them, and rescales them to the same task-balanced prediction magnitude. This removes designed alignment while retaining both forms of heterogeneity.

\begin{table}[!tbp]
\centering
\small
\begin{tabular}{lccc}
\toprule
Method & Excess MSE & Worst-task excess MSE & Task-specific component MSE \\
\midrule
Pool & 0.9359 (0.0256) & 2.8854 (0.5577) & 1.0006 (0.0045) \\
STL & 0.1636 (0.0079) & 0.4031 (0.1017) & 0.1993 (0.0104) \\
HPS & 0.1549 (0.0069) & 0.3467 (0.0634) & 0.1981 (0.0197) \\
MMoE & 0.1797 (0.0087) & 0.4153 (0.0787) & 0.2225 (0.0146) \\
ARMUL & 0.1513 (0.0066) & 0.3394 (0.0604) & 0.1997 (0.0217) \\
FLARCC & 0.2012 (0.0156) & 0.4383 (0.0797) & 0.2666 (0.0293) \\
COVER & 0.1508 (0.0064) & 0.3363 (0.0634) & 0.1936 (0.0195) \\
\bottomrule
\end{tabular}
\caption{Performance under random alignment between covariate and posterior heterogeneity. Entries are means (standard deviations) over 100 repetitions.}
\label{tab:random-alignment}
\end{table}

Table~\ref{tab:random-alignment} shows that \COVER{} remains competitive, with excess error 0.1508 versus 0.1513 for ARMUL and 0.1549 for HPS, and the smallest worst-task error. The margin narrows because overlap matrices are less informative when posterior contrasts have no systematic relation to covariate support.

\subsection{\change{Initialization, Architecture, and Optimization-Budget Control}}\label{sec:additional-sensitivity}

On one moderate joint-heterogeneity data set, we refit HPS and \COVER{} from five initializations under six architectures. Beyond the base widths \((32)\) for \(g\), \((48)\) for \(z\), and \(d=24\), we use \(d=12,36\); narrow widths \((24),(32)\); wide widths \((64),(96)\); and two-layer widths \((32,32),(48,48)\), respectively. \change{Data, optimizer, and validation stopping are unchanged across these fits; all six architectures use the regularization grid \(\{0,0.1,0.3,1,3,10\}\).}

\begin{figure}[!tbp]
\centering
\includegraphics[width=0.96\textwidth]{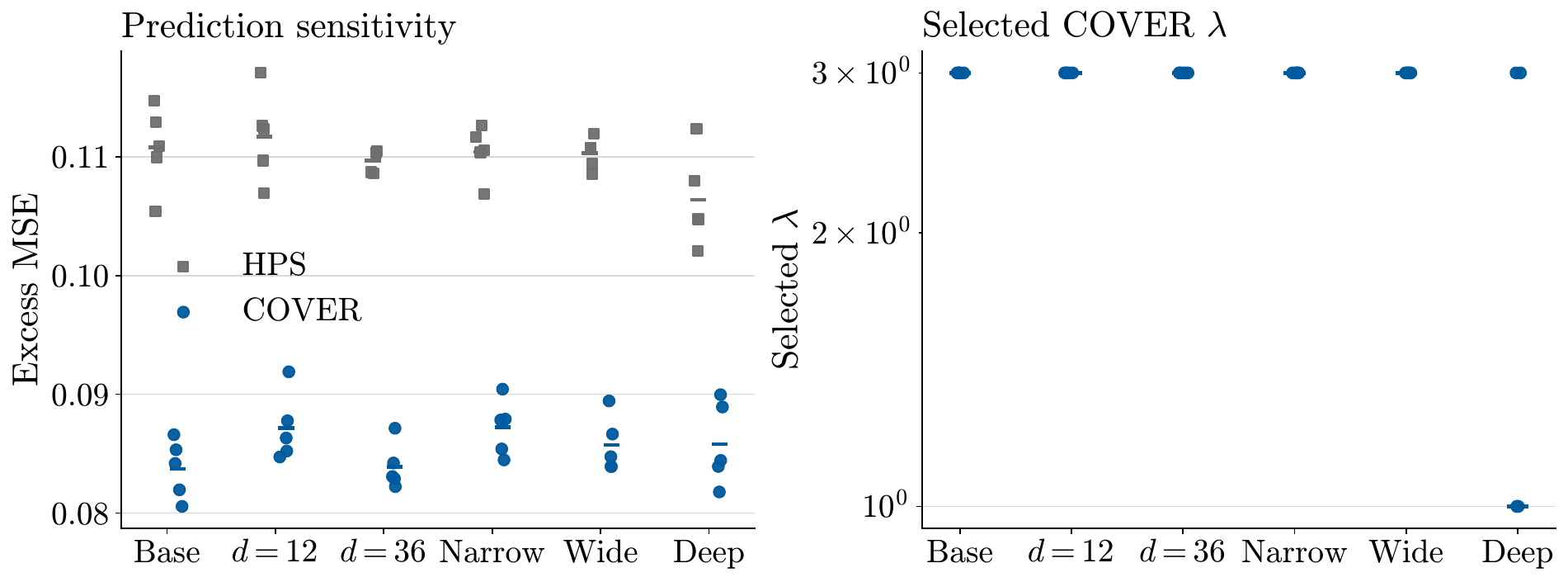}
\caption{Sensitivity to initialization and architecture. Points show the task-balanced excess error from five initializations on a fixed data set; horizontal summaries indicate the corresponding means.}
\label{fig:sensitivity}
\end{figure}

Across all configurations in Figure~\ref{fig:sensitivity}, \COVER{} improves on HPS. The result is not tied to the base width or \(d=p\), and initialization variation is usually smaller than the method gap. This supports prespecifying the main architecture and treating architecture choice as sensitivity analysis.

\change{To separate the covariate-overlap penalty from the additional optimization stage, we conduct a separate 100-repetition diagnostic. The \(\lambda=0\) candidate in the \COVER{} search is the selected HPS checkpoint, whereas Continued-HPS continues that checkpoint under the same 1,000-update limit and validation stopping rule as each positive-\(\lambda\) \COVER{} path, providing a matched continuation-budget control. Mean (SD) excess MSE for HPS, Continued-HPS, and \COVER{} is, respectively, 0.2280 (0.0136), 0.2240 (0.0138), and 0.1022 (0.0061) under homogeneity, and 0.1301 (0.0092), 0.1296 (0.0092), and 0.1059 (0.0066) under strong joint heterogeneity. These summaries come from the separate diagnostic run and therefore differ slightly from Table~\ref{tab:simulation-main}. \COVER{} improves on Continued-HPS in every repetition in both settings, while continuation alone changes HPS only modestly.}

\subsection{\change{Detailed Resource Use}}\label{sec:additional-resources}

Using the strong joint-heterogeneity DGP and full validation workflow, this table reports the largest point on each scaling axis: \(T=192,p=24\) and \(T=48,p=200\). Memory is peak accelerator allocation, predictive parameters include quantities needed at test time, and regularization parameters are medians across 100 repetitions.

\begin{table}[!tbp]
\centering
\scriptsize
\setlength{\tabcolsep}{4pt}
\begin{tabular}{lcccccc}
\toprule
& \multicolumn{3}{c}{\(T=192\)} & \multicolumn{3}{c}{\(p=200\)} \\
\cmidrule(lr){2-4}\cmidrule(lr){5-7}
Method & Memory (MB) & Pred. params. & Selected \(\lambda\) & Memory (MB) & Pred. params. & Selected \(\lambda\) \\
\midrule
Pool & 91.7 & 833 & 0 & 79.9 & 6465 & 0 \\
STL & 77.2 & 159936 & 0 & 84.9 & 310320 & 0 \\
HPS & 82.0 & 7817 & 0 & 80.1 & 18441 & 0 \\
MMoE & 117.4 & 56896 & 0 & 93.1 & 116592 & 0 \\
ARMUL & 82.0 & 7817 & 0.1 & 80.1 & 18441 & 0.3 \\
FLARCC & 82.0 & 7817 & 0.01 & 80.1 & 18441 & 0.3 \\
COVER & 217.1 & 7817 & 3 & 88.0 & 18441 & 10 \\
\bottomrule
\end{tabular}
\caption{Resource use at the largest task-count and input-dimension settings. Memory and predictive parameter counts are means; selected regularization parameters are medians over 100 repetitions.}
\label{tab:scaling-resources}
\end{table}

Table~\ref{tab:scaling-resources} shows that \COVER{} and HPS have equal predictive parameter counts. During optimization, positive-\(\lambda\) \COVER{} also maintains \(\binom{T}{2}d\) auxiliaries: 440,064 coordinates at \(T=192\) and 27,072 at \(T=48\). They are discarded for prediction. We use all pairs to match Section~\ref{sec:estimator}; a prespecified sparse graph can reduce this cost for larger \(T\), but is not evaluated.

\subsection{\change{Overlap-Adaptive Behavior for an Outlying Task}}\label{sec:additional-outlier}

To examine overlap adaptation for an outlying task, we create 24 tasks with \(p=29\), \(d=24\), and one task whose conditional mean is an outlier. The first and second halves of the representation coordinates form two blocks. The 23 inlier tasks have unit variance in the first block and variance \(\nu\in\{0.001,0.003,0.01,0.03,1\}\) in the second; the outlier reverses these variances and has a task-specific component of magnitude 1.5 entirely in the second block. Each task again has 100, 200, and 3,000 observations for training, validation, and testing. We fit \COVER{} to isolate how the learned overlap matrix modulates integration as the outlier's shared covariate support changes.

For \(r=\operatorname{rank}(\overline\Sigma)\), normalize pairwise overlap as \(r^{-1}\operatorname{tr}(\overline\Sigma^\dagger\Omega_{ts})\) and average it over the 23 outlier pairs. The population curve uses true second moments and the learned curve uses training-sample moments, allowing the population and fitted overlap geometries to be compared.

\begin{figure}[!tbp]
\centering
\includegraphics[width=0.97\textwidth]{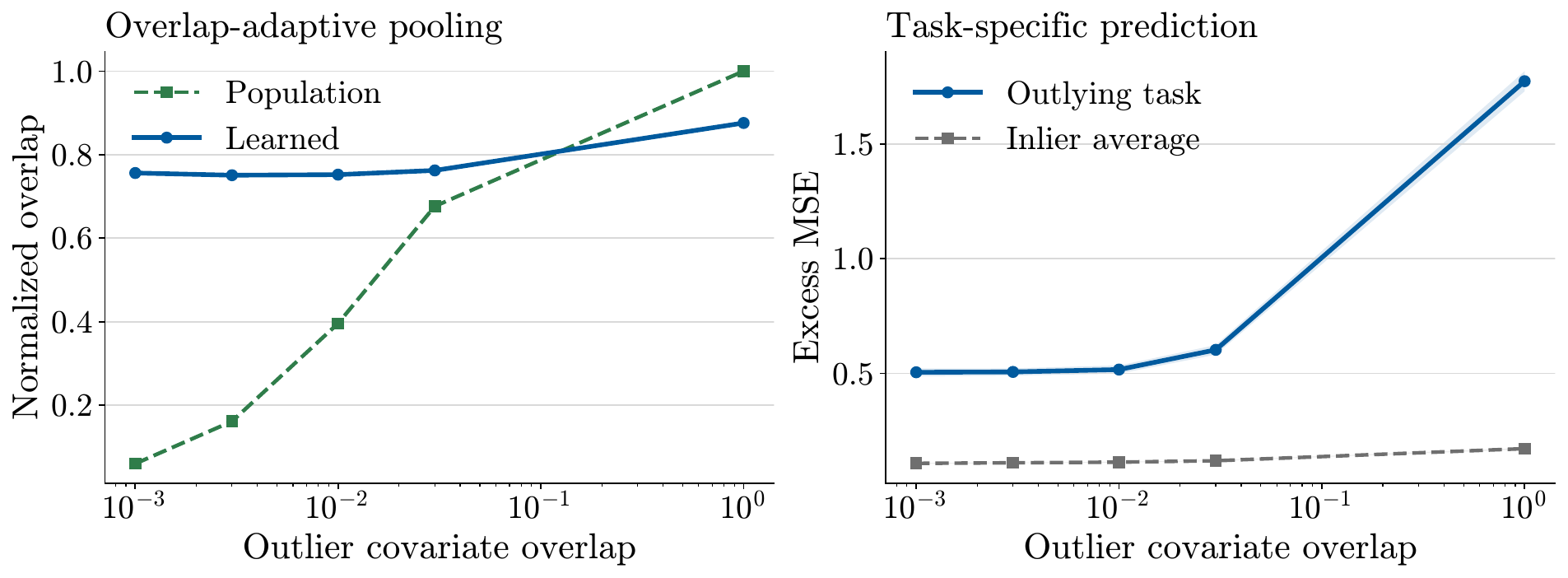}
\caption{Overlap-adaptive behavior for one conditional-mean outlier. The horizontal axis changes its covariate overlap with the remaining tasks. Panels report population and learned normalized overlap relative to the pooled second-moment matrix, together with outlier and inlier prediction errors over 100 repetitions.}
\label{fig:outlier}
\end{figure}

Figure~\ref{fig:outlier} separates the population mechanism from its learned approximation. Population overlap becomes small as shared covariate support disappears, while learned normalized overlap varies more moderately in the lowest-overlap settings. The experiment therefore describes the behavior of the fitted procedure. Across this path, \COVER{} uses the fitted overlap geometry to modulate integration, with its strongest protection against over-pooling occurring when the outlying task has limited covariate overlap with the remaining tasks.

\subsection{\change{Numerical Verification of the Fixed-Representation Identity}}\label{sec:additional-theory}

We verify Theorem~\ref{thm:fixed-bias-variance} for six tasks with a six-dimensional Gaussian fixed representation and \(n=100\) observations per task. The equal design sets every second moment to \(I_6\). The diagonal and nearly singular designs cycle task-specific high-variance coordinates and use low eigenvalues 0.20 and 0.03, respectively. The rotated design uses task-specific orthogonal bases with eigenvalues 1 and 0.15, while the random positive-definite design draws independent bases and eigenvalues uniformly from \([0.08,1.2]\). Centered coefficients are generated once per design. For each of 31 regularization-parameter values, we compare the risk predicted by Theorem~\ref{thm:fixed-bias-variance} with the Monte Carlo risk from 2,000 independent Gaussian noise draws; the reported relative discrepancy is the absolute difference divided by the theoretical risk.

\begin{table}[!tbp]
\centering
\small
\begin{tabular}{lccc}
\toprule
Second-moment design & Values of \(\lambda\) & Median relative error & Maximum relative error \\
\midrule
Diagonal & 31 & 0.00046 & 0.00310 \\
Equal & 31 & 0.00104 & 0.00288 \\
Near-singular & 31 & 0.00278 & 0.00432 \\
Random positive definite & 31 & 0.00140 & 0.00348 \\
Rotated & 31 & 0.00103 & 0.00169 \\
\bottomrule
\end{tabular}
\caption{Numerical verification of the fixed-representation risk identity.}
\label{tab:theory-verification}
\end{table}

The maximum relative discrepancy in Table~\ref{tab:theory-verification} is 0.00432. Agreement across unequal, rotated, and nearly singular designs verifies the constants and generalized eigensystem numerically, but does not replace the proof.

\subsection{\change{Tissue-Specific GTEx Results}}\label{sec:additional-gtex}

\begin{sidewaystable}[p]
\centering
\scriptsize
\setlength{\tabcolsep}{2.7pt}
\begin{tabular}{lccccccc}
\toprule
Tissue & Pool & STL & HPS & MMoE & ARMUL & FLARCC & COVER \\
\midrule
\multicolumn{8}{l}{\textit{JAM2}} \\
Amygdala & 0.3204 (0.0237) & 0.3306 (0.0499) & 0.2803 (0.0403) & 0.3458 (0.0844) & \underline{0.2744 (0.0351)} & 0.2778 (0.0336) & \textbf{0.2718 (0.0371)} \\
Anterior cingulate cortex & 0.4765 (0.0275) & 0.5161 (0.0599) & 0.4219 (0.0439) & 0.5036 (0.0602) & \textbf{0.4070 (0.0390)} & \underline{0.4123 (0.0478)} & 0.4136 (0.0405) \\
Caudate & 0.2592 (0.0121) & 0.2490 (0.0174) & \textbf{0.2172 (0.0149)} & 0.2346 (0.0151) & \underline{0.2191 (0.0149)} & 0.2218 (0.0173) & 0.2197 (0.0196) \\
Cerebellar hemisphere & 0.1805 (0.0159) & \underline{0.1354 (0.0127)} & \textbf{0.1337 (0.0102)} & 0.1368 (0.0122) & 0.1380 (0.0118) & 0.1376 (0.0113) & 0.1382 (0.0119) \\
Cerebellum & 0.2783 (0.0172) & 0.2336 (0.0235) & \underline{0.2236 (0.0218)} & 0.2321 (0.0288) & 0.2254 (0.0200) & 0.2254 (0.0204) & \textbf{0.2204 (0.0171)} \\
Cortex & 0.4499 (0.0304) & 0.4541 (0.0301) & 0.4275 (0.0341) & 0.4483 (0.0386) & \textbf{0.4188 (0.0309)} & 0.4283 (0.0334) & \underline{0.4201 (0.0321)} \\
Frontal cortex & 0.4824 (0.0415) & 0.5140 (0.0283) & 0.4423 (0.0300) & 0.4807 (0.0364) & \underline{0.4353 (0.0280)} & 0.4450 (0.0294) & \textbf{0.4317 (0.0270)} \\
Hypothalamus & 0.3243 (0.0242) & 0.3593 (0.0219) & 0.3244 (0.0296) & 0.3624 (0.0349) & \textbf{0.3144 (0.0273)} & 0.3252 (0.0299) & \underline{0.3151 (0.0254)} \\
Nucleus accumbens & 0.2885 (0.0162) & 0.3292 (0.0161) & 0.2770 (0.0158) & 0.3080 (0.0215) & \underline{0.2703 (0.0134)} & 0.2783 (0.0154) & \textbf{0.2702 (0.0137)} \\
Putamen & 0.2358 (0.0145) & 0.2310 (0.0213) & \textbf{0.2090 (0.0158)} & 0.2175 (0.0226) & \underline{0.2090 (0.0141)} & 0.2121 (0.0143) & 0.2110 (0.0153) \\
Spinal cord & 0.1966 (0.0172) & 0.1896 (0.0137) & \textbf{0.1726 (0.0184)} & 0.1834 (0.0193) & \underline{0.1753 (0.0196)} & 0.1795 (0.0221) & 0.1773 (0.0224) \\
\midrule
\multicolumn{8}{l}{\textit{SH2D2A}} \\
Amygdala & \textbf{1.2427 (0.1972)} & 1.4763 (0.3785) & 1.2786 (0.2559) & 1.3730 (0.2599) & 1.2680 (0.2397) & 1.2700 (0.2439) & \underline{1.2566 (0.2386)} \\
Anterior cingulate cortex & \textbf{0.9910 (0.1798)} & 1.3606 (0.2044) & 1.0381 (0.2044) & 1.3046 (0.2777) & 0.9932 (0.1950) & 1.0246 (0.2077) & \underline{0.9920 (0.1817)} \\
Caudate & \underline{0.5965 (0.0376)} & 0.6706 (0.0370) & 0.6074 (0.0327) & 0.6508 (0.0450) & 0.5993 (0.0385) & 0.6052 (0.0347) & \textbf{0.5949 (0.0317)} \\
Cerebellar hemisphere & \textbf{0.6450 (0.0553)} & 0.7419 (0.0880) & 0.6718 (0.0561) & 0.7181 (0.0640) & \underline{0.6646 (0.0447)} & 0.6682 (0.0503) & 0.6692 (0.0576) \\
Cerebellum & \underline{0.5252 (0.0286)} & 0.5569 (0.0368) & 0.5318 (0.0316) & 0.5508 (0.0369) & \textbf{0.5247 (0.0309)} & 0.5369 (0.0294) & 0.5289 (0.0317) \\
Cortex & \textbf{0.5151 (0.0281)} & 0.5685 (0.0360) & 0.5292 (0.0360) & 0.5421 (0.0352) & 0.5274 (0.0402) & 0.5272 (0.0338) & \underline{0.5252 (0.0346)} \\
Frontal cortex & \underline{0.4767 (0.0366)} & 0.5331 (0.0489) & 0.4818 (0.0368) & 0.5134 (0.0685) & \textbf{0.4740 (0.0321)} & 0.4790 (0.0334) & 0.4802 (0.0375) \\
Hypothalamus & \textbf{0.6259 (0.0735)} & 0.7693 (0.0726) & 0.6600 (0.0829) & 0.7158 (0.0989) & \underline{0.6444 (0.0759)} & 0.6485 (0.0815) & 0.6447 (0.0622) \\
Nucleus accumbens & 0.6801 (0.0507) & 0.7205 (0.0493) & \textbf{0.6539 (0.0413)} & 0.6944 (0.0562) & \underline{0.6558 (0.0440)} & 0.6606 (0.0424) & 0.6578 (0.0435) \\
Putamen & 0.5351 (0.0447) & 0.5960 (0.0668) & 0.5404 (0.0415) & 0.5540 (0.0568) & \underline{0.5338 (0.0505)} & 0.5360 (0.0478) & \textbf{0.5236 (0.0482)} \\
Spinal cord & 0.6317 (0.0448) & 0.6354 (0.0346) & \underline{0.6170 (0.0390)} & 0.6360 (0.0625) & 0.6295 (0.0572) & 0.6194 (0.0339) & \textbf{0.6092 (0.0350)} \\
\bottomrule
\end{tabular}
\caption{Tissue-specific standardized mean squared errors in GTEx. Entries are means (standard deviations) over 20 repeated five-fold cross-validation partitions.}
\label{tab:gtex-tissue}
\end{sidewaystable}

This table disaggregates the GTEx analysis under the same protocol. Within each of 20 repeated five-fold partitions, squared errors are pooled over a tissue's out-of-fold observations; entries summarize these 20 errors.

For Figure~\ref{fig:gtex-overlap}, let \(A_{ts}=(\widehat\Sigma_t+\widehat\Sigma_s)/2\). The fitted matrices are positive definite in this analysis, and each cell is \(d^{-1}\operatorname{tr}(A_{ts}^{-1}\widehat\Omega_{ts})\), the mean generalized eigenvalue of \((\widehat\Omega_{ts},A_{ts})\). The value lies in \([0,1]\), is invariant to a common invertible change of representation coordinates, and measures relative covariate overlap rather than the dimension of a shared subspace.

Table~\ref{tab:gtex-tissue} gives all 22 response--tissue comparisons. No method is uniformly best. \COVER{} improves on HPS in 17, ranks first or second in 12, and is among the top three in 19, showing that the benefits of pooling and overlap-adaptive integration vary across tissues.

\flushbottom

\section{\change{Extension to Multiway Pooling}}\label{app:multiway-pooling}

The pairwise construction in Section~\ref{sec:overlap} extends directly to simultaneous pooling of three or more tasks. For a group \(\mathcal I\subseteq[T]\), define
\begin{equation}
\mathcal C_{\mathcal I}(\{\beta_t\}_{t\in\mathcal I})
=\min_{a\in\mathbb R^d}\sum_{t\in\mathcal I}\|\beta_t-a\|_{\Sigma_t}^2.
\label{eq:multiway-cost}
\end{equation}
Thus every task in the group is moved toward one auxiliary coefficient, with its prediction change measured under its own covariate distribution. Let \(S_{\mathcal I}=\sum_{t\in\mathcal I}\Sigma_t\) and \(b_{\mathcal I}=\sum_{t\in\mathcal I}\Sigma_t\beta_t\). The normal equation is \(S_{\mathcal I}a=b_{\mathcal I}\), which is always solvable because \(b_{\mathcal I}\in\Range(S_{\mathcal I})\). Consequently,
\[
\mathcal C_{\mathcal I}
=\sum_{t\in\mathcal I}\beta_t^\top\Sigma_t\beta_t
-b_{\mathcal I}^\top S_{\mathcal I}^\dagger b_{\mathcal I},
\qquad
a_{\mathcal I}^{\ast}=S_{\mathcal I}^\dagger b_{\mathcal I}+v,
\quad v\in\Null(S_{\mathcal I}).
\]
As with the pairwise auxiliary formulation, \eqref{eq:multiway-cost} can be optimized without forming a pseudoinverse: one introduces one auxiliary coefficient for each selected group and jointly minimizes over that coefficient. Summing these costs over a prespecified collection of groups gives a multiway covariate-overlap penalty. Groups of size three, four, or five can therefore be incorporated when such higher-order structure is scientifically specified.

\noindent\emph{The all-task limit.}
At the other extreme, take \(\mathcal I=[T]\). This gives a single global pooling cost
\[
\mathcal C_{[T]}=\min_{a\in\mathbb R^d}
\sum_{t=1}^T\|\beta_t-a\|_{\Sigma_t}^2,
\]
which moves all task-specific coefficients toward one covariate-weighted center. This is a valid and computationally attractive alternative: it uses only one auxiliary coefficient rather than one for every task pair. It is most natural when the tasks form one coherent group and a single common center adequately describes their posterior heterogeneity.

\noindent\emph{Equivalence under equal second moments.}
If \(\Sigma_t=\Sigma\) for every \(t\in\mathcal I\), then
\[
\mathcal C_{\mathcal I}
=\sum_{t\in\mathcal I}\|\beta_t-\bar\beta_{\mathcal I}\|_\Sigma^2
=\frac1{|\mathcal I|}\sum_{\substack{t<s\\t,s\in\mathcal I}}
\|\beta_t-\beta_s\|_\Sigma^2,
\qquad
\bar\beta_{\mathcal I}=\frac1{|\mathcal I|}\sum_{t\in\mathcal I}\beta_t.
\]
In particular, with equal second moments, all-task pooling and complete pairwise pooling are identical after rescaling the regularization parameter. Higher-order pooling therefore adds no new structure in this setting; the distinction arises only when the task-specific second moments differ.

\noindent\emph{Local versus global dependence.}
With unequal second moments, the all-task center and induced weights depend on every task through \(S_{[T]}\). In one dimension, writing \(\Sigma_t=\sigma_t>0\),
\[
\mathcal C_{[T]}=\frac1{\sum_{\ell=1}^T\sigma_\ell}
\sum_{1\le t<s\le T}\sigma_t\sigma_s(\beta_t-\beta_s)^2,
\qquad
\mathcal C_{ts}=\frac{\sigma_t\sigma_s}{\sigma_t+\sigma_s}(\beta_t-\beta_s)^2.
\]
Thus a third task changes the global weight between \(t\) and \(s\) even if their second moments remain fixed. If \(\sigma_1=\sigma_2=1\) and \(\sigma_3=M\), the global coefficient of \((\beta_1-\beta_2)^2\) is \(1/(M+2)\), whereas its pairwise coefficient remains \(1/2\). More generally,
\[
\frac1{T-1}\sum_{1\le t<s\le T}\mathcal C_{ts}
\le \mathcal C_{[T]}.
\]
Evaluating every pairwise problem at the all-task minimizer counts each task cost \(T-1\) times. This records the flexibility of pair-specific centers, not a prediction-risk ordering.

\noindent\emph{Why the main method uses pairs.}
The pairwise cost is the smallest nontrivial pooling unit and has the exact support interpretation
\(\Range(\Omega_{ts})=\Range(\Sigma_t)\cap\Range(\Sigma_s)\): the task-specific coefficients are pooled only along directions supported by both covariate distributions. Its value depends only on those two tasks, so adding a new task creates new pairs without redefining the existing unscaled pairwise costs. This local property is lost under a single all-task center and is useful when the tasks contain clusters or have uneven covariate overlap. Pairwise pooling also avoids selecting higher-order groups, whereas using every possible group would introduce exponentially many terms. These considerations motivate the pairwise criterion in the main method. The all-task version remains a useful lower-cost alternative, and selected multiway groups can be added when substantive knowledge supports them.

\section{\change{Supplementary Theory for Section~\ref{sec:theory}}}
\label{app:supp-theory}
\label{app:end-to-end-lemmas}

This appendix gives the complexity measures and auxiliary results for Section~\ref{sec:theory}; their proofs appear in Section~\ref{app:supplementary-proofs}.

\subsection{\change{Complexity Measures}}\label{app:complexity-notation}

Let \(\{\epsilon_{ti}\}_{t\in[T],i\in[n]}\) be independent Rademacher variables, independent of the observations. The task-balanced prediction and taskwise coordinate-product complexities are
\[
\begin{aligned}
\mathscr R_N(\mathcal F)
&=\mathbb E\sup_{h\in\mathcal H}
\left|\frac1N\sum_{t=1}^T\sum_{i=1}^n\epsilon_{ti}f_{h,t}(X_{ti})\right|,\\
\widetilde{\mathscr R}_n(\mathcal Z)
&=\max_{\substack{t\in[T]\\j,k\in[d]}}
\mathbb E\sup_{z\in\mathcal Z}
\left|\frac1n\sum_{i=1}^n\epsilon_{ti}z_j(X_{ti})z_k(X_{ti})\right|.
\end{aligned}
\]
Both expectations include covariates and Rademacher variables. Product-class contraction bounds the second quantity by \(B_z\) times the coordinatewise representation complexities.

For the localized analysis, let \(\gamma_{ti}\) be independent standard Gaussian variables and \(\eta_{ti}\) independent \(d\)-dimensional standard Gaussian vectors. The worst-case Gaussian complexities are
\[
\begin{aligned}
\mathscr G_N(\mathcal G)
&=\sup_{\{x_{ti}\}_{t\in[T],i\in[n]}}\mathbb E_\gamma\sup_{g\in\mathcal G}
\left|\frac1N\sum_{t=1}^T\sum_{i=1}^n\gamma_{ti}g(x_{ti})\right|,\\
\mathscr G_N(\mathcal Z)
&=\sup_{\{x_{ti}\}_{t\in[T],i\in[n]}}\mathbb E_\eta\sup_{z\in\mathcal Z}
\left|\frac1N\sum_{t=1}^T\sum_{i=1}^n\eta_{ti}^{\top}z(x_{ti})\right|.
\end{aligned}
\]
where each supremum ranges over admissible task-indexed covariate arrays \(\{x_{ti}\}_{t\in[T],i\in[n]}\). Finally, writing \(\ell_{h,t}(x,y)=\{y-f_{h,t}(x)\}^2/2\), define
\[
\mathscr R_N^{\mathrm{loc}}(\rho)
=\mathbb E\sup_{\substack{h\in\mathcal H\\\mathcal R(h)\le\rho}}
\left|\frac1N\sum_{t=1}^T\sum_{i=1}^n\epsilon_{ti}\ell_{h,t}(X_{ti},Y_{ti})\right|,
\qquad \rho>0.
\]
For a matrix \(A\), \(\|A\|_{\mathrm{op}}\) denotes its operator norm.

\subsection{\change{Estimation and Stability of the Overlap Matrices}}

\begin{lemma}[Uniform second-moment estimation]\label{lem:second-moment}
Under Assumption~\ref{ass:neural-classes}, with probability at least \(1-\delta\),
\[
\sup_{z\in\mathcal Z}\max_{t\in[T]}
\|\widehat\Sigma_t(z)-\Sigma_t(z)\|_{\mathrm{op}}
\le r_{\Sigma,n}(\delta).
\]
\end{lemma}

\begin{lemma}[Stability of overlap matrices]\label{lem:overlap-stability}
Suppose Assumption~\ref{ass:uniform-spectrum} holds and
\[
\Delta_\Sigma
=\sup_{z\in\mathcal Z}\max_{t\in[T]}
\|\widehat\Sigma_t(z)-\Sigma_t(z)\|_{\mathrm{op}}
\le\frac{m_z}{2}.
\]
Then
\[
\sup_{z\in\mathcal Z}\max_{1\le t<s\le T}
\|\widehat\Omega_{ts}(z)-\Omega_{ts}(z)\|_{\mathrm{op}}
\le
7\left(\frac{M_z}{m_z}\right)^2
\Delta_\Sigma.
\]
\end{lemma}

\subsection{\change{Prediction Complexity and Explicit Neural-Network Rates}}\label{app:local-analysis}

\begin{proposition}[Localized prediction complexity]\label{prop:local-complexity}
Under Assumption~\ref{ass:neural-classes}, a universal constant \(C>0\) satisfies \(\mathscr R_N^{\mathrm{loc}}(\rho)\le C\{\sqrt{r_N\rho}+r_N\}\) for every \(\rho>0\). Hence the localized fixed-point scale is at most \(Cr_N\).
\end{proposition}

For an explicit specialization, suppose \(\|X\|_2\le B_X\), \(g\) is a depth-\(L_g\) ReLU network, and each coordinate of \(z\) is a depth-\(L_z\) ReLU network. Network bias parameters, when present, are included in the layerwise norm bounds through the usual homogeneous-coordinate representation, with \(B_X\) interpreted as a bound on the augmented input norm. If the layerwise Frobenius-norm bounds are \(F_{g,\ell}\) and \(F_{z,\ell}\), respectively, define
\[
\Gamma_g
=B_X\left(\prod_{\ell=1}^{L_g}F_{g,\ell}\right)
\sqrt{L_g+1+\log p},
\qquad
\Gamma_z
=B_X\left(\prod_{\ell=1}^{L_z}F_{z,\ell}\right)
\sqrt{L_z+1+\log p}.
\]

\begin{corollary}[Explicit neural-network rates]\label{cor:explicit-neural-rate}
Under Assumptions~\ref{ass:neural-classes} and \ref{ass:uniform-spectrum} and the preceding norm constraints, there is a universal constant \(C>0\) such that
\begin{align}
\sqrt{r_N}
&\le C\left[
\left\{
\frac{\Gamma_g+B_\beta d\Gamma_z}{\sqrt{Tn}}
+B_zB_\beta\sqrt{\frac dn}
\right\}\log^2(eTn)
+\frac{B\log(eTn)}{\sqrt{Tn}}
\right],
\label{eq:explicit-local-scale}\\
\widetilde{\mathscr R}_n(\mathcal Z)
&\le \frac{CB_z\Gamma_z}{\sqrt n},
\qquad
r_{\Sigma,n}(\delta)
\le
\frac{CdB_z\Gamma_z}{\sqrt n}
+dB_z^2\sqrt{\frac{2\log(Td^2/\delta)}n}.
\label{eq:explicit-moment-rate}
\end{align}
Substitution of \eqref{eq:explicit-local-scale}--\eqref{eq:explicit-moment-rate} into Theorem~\ref{thm:optimistic-oracle} gives a fully explicit rate.
\end{corollary}

The first fraction in \eqref{eq:explicit-local-scale} is learned at the combined sample size \(Tn\), while the coefficient term is learned at the within-task sample size \(n\). Equation~\eqref{eq:explicit-moment-rate} gives the taskwise scale for estimating the overlap matrices.

\begin{lemma}[Localized relative deviation]\label{lem:relative-deviation}
Under Assumption~\ref{ass:neural-classes}, for every
\(0<\delta<1\), with probability at least \(1-\delta\), simultaneously
for every \(h\in\mathcal H\) and every \(K\ge2\),
\[
\mathcal R(h)
\le
\frac{K}{K-1}\widehat{\mathcal R}(h)
+CK\left\{
r_N+\frac{B^2\log(1/\delta)}N
\right\},
\]
where \(C>0\) is a universal constant.
\end{lemma}

\subsection{\change{Proofs of the Supplementary Results}}
\label{app:supplementary-proofs}

\begin{proofof}{Lemma~\ref{lem:second-moment}}
Fix a task \(t\) and coordinates \(j,k\), and set
\(\phi_z(x)=z_j(x)z_k(x)\). Introduce an independent ghost sample
\(\{X_{ti}'\}_{i=1}^n\) from \(\mathbb P_t^X\). Jensen's inequality, followed by Rademacher symmetrization, gives
\[
\begin{aligned}
\mathbb E\sup_{z\in\mathcal Z}
\left|
\frac1n\sum_{i=1}^n\phi_z(X_{ti})
-\mathbb E_t\phi_z(X)
\right|
&\le
\mathbb E\sup_{z\in\mathcal Z}
\left|
\frac1n\sum_{i=1}^n
\{\phi_z(X_{ti})-\phi_z(X_{ti}')\}
\right|\\
&=
\mathbb E\sup_{z\in\mathcal Z}
\left|
\frac1n\sum_{i=1}^n
\epsilon_{ti}\{\phi_z(X_{ti})-\phi_z(X_{ti}')\}
\right|\\
&\le
2\mathbb E\sup_{z\in\mathcal Z}
\left|
\frac1n\sum_{i=1}^n
\epsilon_{ti}\phi_z(X_{ti})
\right|\\
&\le2\widetilde{\mathscr R}_n(\mathcal Z).
\end{aligned}
\]
The equality uses the exchangeability of every pair
\((X_{ti},X_{ti}')\), and the next inequality follows from the triangle inequality and the identical distribution of the original and ghost samples.

Let
\[
D_{tjk}
=\sup_{z\in\mathcal Z}
\left|
\frac1n\sum_{i=1}^n\phi_z(X_{ti})
-\mathbb E_t\phi_z(X)
\right|.
\]
Assumption~\ref{ass:neural-classes} implies
\(|\phi_z(x)|\le B_z^2\). Replacing one covariate can change
\(D_{tjk}\) by at most \(2B_z^2/n\). McDiarmid's inequality therefore yields, for every \(u>0\),
\[
\Pr\{D_{tjk}-\mathbb ED_{tjk}>u\}
\le
\exp\left\{-\frac{nu^2}{2B_z^4}\right\}.
\]
Taking \(u=B_z^2\{2\log(Td^2/\delta)/n\}^{1/2}\) makes the right-hand side
\(\delta/(Td^2)\). A union bound over all \(T d^2\) choices of
\((t,j,k)\) then shows that, with probability at least \(1-\delta\),
\[
\sup_{z\in\mathcal Z}
\max_{\substack{t\in[T]\\j,k\in[d]}}
\left|
\{\widehat\Sigma_t(z)-\Sigma_t(z)\}_{jk}
\right|
\le
2\widetilde{\mathscr R}_n(\mathcal Z)
+B_z^2\sqrt{\frac{2\log(Td^2/\delta)}{n}}.
\]
To pass from entrywise to operator-norm control, let \(A\) be any
\(d\times d\) matrix and let \(u,v\) be unit vectors. Then
\[
|u^\top Av|
\le
\max_{j,k}|A_{jk}|\,\|u\|_1\|v\|_1
\le d\max_{j,k}|A_{jk}|,
\]
where the last step uses
\(\|u\|_1,\|v\|_1\le\sqrt d\).
Taking the supremum over unit vectors and applying the resulting inequality to every second-moment difference proves the claim.
\end{proofof}

\begin{proofof}{Lemma~\ref{lem:overlap-stability}}
Fix \(z\) and a task pair, and abbreviate the corresponding population matrices by \(A\) and \(B\) and their empirical versions by \(\widehat A\) and \(\widehat B\). Let
\(S=A+B\) and \(\widehat S=\widehat A+\widehat B\). Assumption~\ref{ass:uniform-spectrum} gives
\(\|A\|_{\mathrm{op}},\|B\|_{\mathrm{op}}\le M_z\) and
\(\|S^{-1}\|_{\mathrm{op}}\le(2m_z)^{-1}\).
On the event in the statement,
\(\widehat A,\widehat B\succeq(m_z/2)I_d\), so
\[
\|\widehat S^{-1}\|_{\mathrm{op}}\le\frac1{m_z},
\qquad
\|\widehat A\|_{\mathrm{op}},
\|\widehat B\|_{\mathrm{op}}
\le M_z+\Delta_\Sigma
\le\frac32M_z.
\]
The resolvent identity
\(\widehat S^{-1}-S^{-1}
=\widehat S^{-1}(S-\widehat S)S^{-1}\) further gives
\[
\|\widehat S^{-1}-S^{-1}\|_{\mathrm{op}}
\le
\frac1{m_z}\,(2\Delta_\Sigma)\,
\frac1{2m_z}
=\frac{\Delta_\Sigma}{m_z^2}.
\]

All four second-moment matrices are positive definite on this event, so their overlap matrices can be written as
\(\Omega=2AS^{-1}B\) and
\(\widehat\Omega=2\widehat A\widehat S^{-1}\widehat B\).
Adding and subtracting intermediate terms yields
\[
\begin{aligned}
\frac12(\widehat\Omega-\Omega)
={}&(\widehat A-A)\widehat S^{-1}\widehat B
+A(\widehat S^{-1}-S^{-1})\widehat B\\
&+AS^{-1}(\widehat B-B).
\end{aligned}
\]
Submultiplicativity of the operator norm bounds the three terms, after restoring the factor two, by
\[
\begin{aligned}
2\|\widehat A-A\|_{\mathrm{op}}
\|\widehat S^{-1}\|_{\mathrm{op}}
\|\widehat B\|_{\mathrm{op}}
&\le \frac{3M_z}{m_z}\Delta_\Sigma,\\
2\|A\|_{\mathrm{op}}
\|\widehat S^{-1}-S^{-1}\|_{\mathrm{op}}
\|\widehat B\|_{\mathrm{op}}
&\le \frac{3M_z^2}{m_z^2}\Delta_\Sigma,\\
2\|A\|_{\mathrm{op}}
\|S^{-1}\|_{\mathrm{op}}
\|\widehat B-B\|_{\mathrm{op}}
&\le \frac{M_z}{m_z}\Delta_\Sigma.
\end{aligned}
\]
Adding these inequalities gives
\[
\|\widehat\Omega-\Omega\|_{\mathrm{op}}
\le
\left\{
\frac{4M_z}{m_z}
+\frac{3M_z^2}{m_z^2}
\right\}\Delta_\Sigma
\le
7\left(\frac{M_z}{m_z}\right)^2\Delta_\Sigma,
\]
where the final inequality uses \(M_z\ge m_z\). The bound is uniform because the event and the spectral constants are uniform over \(z\) and the tasks.
\end{proofof}

\begin{proofof}{Proposition~\ref{prop:local-complexity}}
For this proof only, write \(b=B^2/2\) and
\(D_{\mathcal F}=2(B_g+B_zB_\beta)\). We first establish a Gaussian
comparison for the predictor class, writing
\(\mathscr G_N(\mathcal F)\) for the scalar Gaussian complexity defined
as in Appendix~\ref{app:complexity-notation}, with \(f_{h,t}(x_{ti})\) in place of
\(g(x_{ti})\). The contribution of the common component function is additive,
so the triangle inequality gives
\[
\mathscr G_N(\mathcal F)
\le
\mathscr G_N(\mathcal G)+\mathscr G_N(\mathcal V),
\qquad
\mathcal V
=
\{(x,t)\mapsto z(x)^\top\beta_t:
z\in\mathcal Z,\ \max_{t\in[T]}\|\beta_t\|_2\le B_\beta\}.
\]
Dropping the centering constraint \(\sum_{t=1}^T\beta_t=0\) only enlarges
\(\mathcal V\) and therefore gives a valid upper bound.

Fix an admissible covariate array
\(\{x_{ti}:t\in[T],i\in[n]\}\). For
\(\vartheta=(z,\beta_1,\ldots,\beta_T)\), define the centered Gaussian
process
\[
Z_\vartheta
=\frac1N\sum_{t=1}^T\sum_{i=1}^n
\gamma_{ti}z(x_{ti})^\top\beta_t,
\]
where the \(\gamma_{ti}\) are independent standard Gaussian variables.
For \(\vartheta'=(z',\beta_1',\ldots,\beta_T')\), add and subtract
\(z'(x_{ti})^\top\beta_t\) and use
\((a+b)^2\le2a^2+2b^2\). Since
\(\|\beta_t\|_2\le B_\beta\) and
\(\|z'(x)\|_2\le B_z\),
\[
\mathbb E_\gamma(Z_\vartheta-Z_{\vartheta'})^2
\le
\frac{2B_\beta^2}{N^2}
\sum_{t=1}^T\sum_{i=1}^n\|z(x_{ti})-z'(x_{ti})\|_2^2
+\frac{2B_z^2}{N^2}
\sum_{t=1}^T\sum_{i=1}^n\|\beta_t-\beta_t'\|_2^2.
\]
Let \(\eta_{ti}\in\mathbb R^d\) and
\(\xi_{ti}\in\mathbb R^d\) be mutually independent standard Gaussian
vectors and consider the comparison process whose value at \(\vartheta\) is
\[
\frac{\sqrt2B_\beta}{N}\sum_{t=1}^T\sum_{i=1}^n\eta_{ti}^\top z(x_{ti})
+\frac{\sqrt2B_z}{N}\sum_{t=1}^T\sum_{i=1}^n\xi_{ti}^\top\beta_t.
\]
Its increment variance is the right-hand side of the preceding
inequality. The Gaussian comparison inequality therefore yields,
up to a universal numerical constant,
\[
\mathbb E_\gamma\sup_\vartheta|Z_\vartheta|
\le
C B_\beta\mathscr G_N(\mathcal Z)
+\frac{CB_z}{N}
\mathbb E_\xi
\sup_{\max_{t\in[T]}\|\beta_t\|_2\le B_\beta}
\sum_{t=1}^T
\left(\sum_{i=1}^n\xi_{ti}\right)^\top\beta_t.
\]
For the second term, Cauchy--Schwarz and Jensen's inequality give
\[
\begin{aligned}
\frac{B_z}{N}
\mathbb E_\xi
\sup_{\max_{t\in[T]}\|\beta_t\|_2\le B_\beta}
\sum_{t=1}^T\left(\sum_{i=1}^n\xi_{ti}\right)^\top\beta_t
&\le
\frac{B_zB_\beta}{N}
\sum_{t=1}^T
\mathbb E_\xi\left\|\sum_{i=1}^n\xi_{ti}\right\|_2\\
&\le
\frac{B_zB_\beta}{Tn}\,T\sqrt{nd}
=B_zB_\beta\sqrt{\frac dn}.
\end{aligned}
\]
Taking the supremum over admissible arrays \(\{x_{ti}\}\) gives the intermediate bound
\[
\mathscr G_N(\mathcal F)
\le C\left\{
\mathscr G_N(\mathcal G)
+B_\beta\mathscr G_N(\mathcal Z)
+B_zB_\beta\sqrt{\frac dn}
\right\}.
\]
This calculation identifies why the common component function, shared representation, and
task-specific coefficients have different sample-size scales.

We next localize the squared-loss class. The loss is nonnegative,
bounded by \(b\), and, as a function of the prediction, has
one-Lipschitz derivative. Moreover, the empirical \(L_2\) diameter of
the predictor class is bounded by \(D_{\mathcal F}\). Define, within
this proof,
\[
a_N=
\mathscr G_N(\mathcal F)\log^2(eN)
+\frac{D_{\mathcal F}\log(eN)+\sqrt b}{\sqrt N}
.
\]
Smoothness and nonnegativity give the self-bounding relation
\(\{\partial\ell_{h,t}/\partial f\}^2\le2\ell_{h,t}\), which restricts
the empirical local prediction diameter to order \(\sqrt{\rho}\).
Applying Dudley's entropy integral and then Sudakov comparison yields
an empirical local loss complexity of order \(a_N\sqrt{\rho}\); the
two entropy comparisons account for \(\log^2(eN)\). This is the
smooth nonnegative-loss argument of \citet{srebro2010smoothness}, in
the Gaussian-width form used for multi-task representation classes by
\citet{watkins2024optimistic}.

The class in the proposition is localized by population rather than
empirical risk. We therefore use the standard critical-radius transfer
for bounded nonnegative smooth losses. Specifically, Lemma~2.2 and the
proof of Theorem~1 in \citet{srebro2010smoothness} convert an
empirical-ball bound of order \(a_N\sqrt r\) into a population-ball
bound by applying the local Rademacher argument of Theorem~3.3 and
Corollary~3.5 in \citet{bartlett2005local}. Applied at the critical
radius and then on dyadic population-risk shells, this argument gives
the sub-root majorant
\[
\mathscr R_N^{\mathrm{loc}}(\rho)
\le C\{a_N\sqrt{\rho}+a_N^2\}.
\]
The additive \(a_N^2\) is the lower-scale term introduced by this
transfer; it cannot generally be omitted for a population-localized
class. This transfer is not an invocation of
Lemma~\ref{lem:relative-deviation} below: it is the preceding standard
localization result used to construct the sub-root majorant, whereas
Lemma~\ref{lem:relative-deviation} subsequently uses that majorant to
obtain a uniform risk comparison. The symmetrization, Bernstein
concentration, and peeling steps require independence but not identical
distributions, because every empirical term is paired with its
expectation under the same task and all task weights are equal.

The preceding Gaussian comparison and the definition of \(r_N\) imply
\(a_N\le C\sqrt{r_N}\). Hence
\(\mathscr R_N^{\mathrm{loc}}(\rho)
\le C\{\sqrt{r_N\rho}+r_N\}\). Finally, the fixed-point equation for
the majorant has the form \(\rho=C\sqrt{r_N\rho}+Cr_N\). Setting
\(q=\sqrt{\rho/r_N}\) reduces it to \(q^2=Cq+C\), whose positive
solution is a numerical constant. Thus the positive fixed point is at
most \(Cr_N\), as claimed.
\end{proofof}

\begin{proofof}{Corollary~\ref{cor:explicit-neural-rate}}
Repeat the layerwise contraction-and-duality argument underlying the
norm-based neural-network bound of \citet{golowich2018size}, now with
independent standard Gaussian multipliers. Gaussian contraction passes
through each one-Lipschitz ReLU activation, and Frobenius duality controls
each linear layer by its prescribed norm. Iterating over the layers gives
\[
\mathscr G_N(\mathcal G)
\le\frac{C\Gamma_g}{\sqrt N}.
\]
Applying the same Gaussian argument to each output coordinate of \(z\),
followed by the triangle inequality for the vector Gaussian process, gives
\[
\mathscr G_N(\mathcal Z)
\le\frac{Cd\Gamma_z}{\sqrt N}.
\]
These bounds are valid for the shared network evaluated on the pooled
task-indexed covariate array; the observations need not share a common
covariate distribution. Substitution into \eqref{eq:local-scale}, with
\(N=Tn\), proves \eqref{eq:explicit-local-scale}.

It remains to control the coordinate-product class. For fixed
\(j,k\), both coordinate classes are uniformly bounded by \(B_z\).
The product-class contraction inequality therefore gives
\[
\mathbb E\sup_{z\in\mathcal Z}
\left|
\frac1n\sum_{i=1}^n
\epsilon_{ti}z_j(X_{ti})z_k(X_{ti})
\right|
\le
B_z\{\mathscr R_n(\mathcal Z_j)+\mathscr R_n(\mathcal Z_k)\},
\]
where \(\mathcal Z_j=\{x\mapsto z_j(x):z\in\mathcal Z\}\).
The scalar norm-based bound implies
\(\mathscr R_n(\mathcal Z_j)\le C\Gamma_z/\sqrt n\), uniformly in
\(j\) and \(t\). Maximizing over \(t,j,k\) proves
\(\widetilde{\mathscr R}_n(\mathcal Z)\le
CB_z\Gamma_z/\sqrt n\). Inserting this inequality into
\eqref{eq:covariance-rate} proves \eqref{eq:explicit-moment-rate}.
The fixed-point statement follows from
Proposition~\ref{prop:local-complexity}.
\end{proofof}

\begin{proofof}{Lemma~\ref{lem:relative-deviation}}
We give the reduction to the standard local Rademacher fixed-point
argument. Set \(b=B^2/2\). For \(r>0\), consider the loss functions indexed by
\(\{h\in\mathcal H:\mathcal R(h)\le r\}\).
Because \(0\le\ell_{h,t}\le b\),
\[
\frac1T\sum_{t=1}^T
\mathbb E_t\ell_{h,t}^2
\le b\mathcal R(h).
\]
Symmetrization followed by the bounded empirical-process inequality
therefore shows that, for a fixed radius \(r\), the upper deviation
over this subclass is bounded by
\[
C\left\{
\mathscr R_N^{\mathrm{loc}}(r)
+\sqrt{\frac{br\log(1/\delta)}N}
+\frac{b\log(1/\delta)}N
\right\}.
\]
By Proposition~\ref{prop:local-complexity},
\(C\{\sqrt{r_Nr}+r_N\}\) is a sub-root upper bound whose fixed point is at
most \(Cr_N\). Applying the preceding inequality on geometrically
increasing risk shells and using this fixed-point bound, the usual
peeling argument for local Rademacher classes
\citep{bartlett2005local} consequently
gives, simultaneously for every \(h\in\mathcal H\),
\[
\mathcal R(h)-\widehat{\mathcal R}(h)
\le
C\left[
\sqrt{\mathcal R(h)}
\left\{
\sqrt{r_N}
+\sqrt{\frac{b\log(1/\delta)}N}
\right\}
+r_N
+\frac{b\log(1/\delta)}N
\right].
\]
The same statement is immediate when
\(\mathcal R(h)\le Cr_N\) after increasing \(C\).
Let
\(s=r_N+B^2\log(1/\delta)/N\). Since \(b\le B^2\), the preceding display implies
\(\mathcal R(h)\le\widehat{\mathcal R}(h)
+C\sqrt{\mathcal R(h)s}+Cs\).
For any \(K\ge2\), Young's inequality gives
\(C\sqrt{\mathcal R(h)s}
\le\mathcal R(h)/K+C'Ks\).
After moving \(\mathcal R(h)/K\) to the left, use
\(K/(K-1)\le2\) and \(K^2/(K-1)\le2K\). Enlarging the universal
constant then yields the claimed relative-deviation inequality.
\end{proofof}

\section{\change{Transfer to a New Task}}\label{app:new-task-transfer}

This appendix extends a fitted \COVER{} model to a new but related task. We first give the adaptation rule, then establish fixed-representation and end-to-end guarantees for the resulting predictor.

\subsection{\change{Adaptation Rule}}

The overlap matrices also provide a direct way to adapt a fitted \COVER{}
model to a new task. Suppose \(\widehat g\), \(\widehat z\),
\(\{\widehat\beta_t\}_{t=1}^T\), and
\(\{\widehat\Sigma_t\}_{t=1}^T\) have been learned from the original
tasks, with \(\sum_{t=1}^T\widehat\beta_t=0\). Thus \(\widehat g\) is
the common component function for these \(T\) tasks. A new task, indexed
by 0, supplies i.i.d. observations
\(\{(X_{0i},Y_{0i})\}_{i=1}^{n_0}\) from \(\mathbb P_0\), independently
of the original task data and fitted model. During
adaptation, we hold the original fit fixed and use \(\widehat g\) as the
reference function; no centering constraint involving the new coefficient
is imposed.

Form the new-task second-moment and overlap matrices as
\[
\widehat\Sigma_0=\frac1{n_0}\sum_{i=1}^{n_0}
\widehat z(X_{0i})\widehat z(X_{0i})^\top,
\qquad
\widehat\Omega_{0t}
=2\widehat\Sigma_0(\widehat\Sigma_0+\widehat\Sigma_t)^\dagger
\widehat\Sigma_t.
\]
For \(\lambda_0\ge0\), the adapted coefficient is
\begin{equation}\label{eq:new-task-estimator}
\widehat\beta_0\in
\arg\min_{\|\beta\|_2\le B_\beta}
\left[
\frac1{2n_0}\sum_{i=1}^{n_0}
\{Y_{0i}-\widehat g(X_{0i})-\widehat z(X_{0i})^\top\beta\}^2
+\frac{\lambda_0}{2T}\sum_{t=1}^T
\|\beta-\widehat\beta_t\|_{\widehat\Omega_{0t}}^2
\right].
\end{equation}
The capacity bound is used for the end-to-end result below and may be
chosen large enough to be inactive. Each original coefficient then
influences the new task only along representation directions supported by
both tasks. The regularization parameter \(\lambda_0\) can be selected
using new-task validation data.

To make the update explicit, define
\[
\widehat\Lambda_0=\frac1T\sum_{t=1}^T\widehat\Omega_{0t},
\qquad
\widetilde\beta_0=\widehat\Lambda_0^\dagger
\left(\frac1T\sum_{t=1}^T
\widehat\Omega_{0t}\widehat\beta_t\right),
\qquad
\widehat s_0=\frac1{n_0}\sum_{i=1}^{n_0}
\widehat z(X_{0i})\{Y_{0i}-\widehat g(X_{0i})\}.
\]
The vector \(T^{-1}\sum_{t=1}^T
\widehat\Omega_{0t}\widehat\beta_t\) belongs to
\(\sum_{t=1}^T\Range(\widehat\Omega_{0t})=
\Range(\widehat\Lambda_0)\). Consequently,
\(\widehat\Lambda_0\widetilde\beta_0=
T^{-1}\sum_{t=1}^T\widehat\Omega_{0t}\widehat\beta_t\).
Up to a constant independent of \(\beta\), the transfer penalty in
\eqref{eq:new-task-estimator} is
\(\lambda_0\|\beta-\widetilde\beta_0\|_{\widehat\Lambda_0}^2/2\).
When the capacity bound is inactive, the minimum-norm solution is
\begin{equation}\label{eq:new-task-closed-form}
\widehat\beta_0
=(\widehat\Sigma_0+\lambda_0\widehat\Lambda_0)^\dagger
(\widehat s_0+\lambda_0\widehat\Lambda_0\widetilde\beta_0),
\qquad
\widehat f_0(x)=\widehat g(x)+\widehat z(x)^\top\widehat\beta_0.
\end{equation}

The function \(\widehat g\) in \eqref{eq:new-task-estimator} remains the
center of the original \(T\) tasks; it is not the center of the expanded
collection. If a task-balanced decomposition of all \(T+1\) fitted
functions is desired, it is obtained after adaptation by the exact
recentering
\[
\begin{aligned}
\widehat g^{+}(x)
&=\widehat g(x)+\frac1{T+1}\widehat z(x)^\top\widehat\beta_0,\\
\widehat\beta_t^{+}
&=\widehat\beta_t-\frac{\widehat\beta_0}{T+1},\quad t\in[T],
\qquad
\widehat\beta_0^{+}=\frac{T}{T+1}\widehat\beta_0.
\end{aligned}
\]
Then \(\sum_{t=0}^T\widehat\beta_t^{+}=0\),
\(\widehat g^{+}(x)=(T+1)^{-1}\sum_{t=0}^T\widehat f_t(x)\), and every fitted
function is unchanged. All coefficient differences are also unchanged, so
the recentering does not alter the covariate-overlap penalty. Thus the
original-task-centered coordinates are used only to compute the transfer update,
whereas \(\widehat g^{+}\) is the common component function for the
expanded collection. This is a post-hoc algebraic recentering;
\(\widehat g^{+}\) need not belong to the original class \(\mathcal G\)
unless that class is closed under the displayed update.

By Proposition~\ref{prop:pooling},
\(\Range(\widehat\Omega_{0t})=\Range(\widehat\Sigma_0)\cap
\Range(\widehat\Sigma_t)\), so
\(\Range(\widehat\Lambda_0)\subseteq\Range(\widehat\Sigma_0)\).
The transfer penalty therefore introduces no fitted component in a
direction unsupported by the new task. This property also ensures that
\eqref{eq:new-task-closed-form} solves the normal equation when the
empirical second-moment matrices are singular.

\subsection{\change{Theoretical Guarantees}}

We first isolate the prediction bias--variance tradeoff. Condition on the
original fit and the new-task covariates, suppose the capacity bound is
inactive and \(\widehat\Sigma_0\succ0\), and assume
\[
Y_{0i}=\widehat g(X_{0i})+
\widehat z(X_{0i})^\top\beta_0^\ast+\varepsilon_{0i},
\]
where the errors are independent with conditional mean zero and conditional
variance \(\sigma_0^2\). Let \(\gamma_1,\ldots,\gamma_d\) and
\(u_1,\ldots,u_d\) be the eigenvalues and orthonormal eigenvectors of
\(\widehat\Sigma_0^{-1/2}\widehat\Lambda_0
\widehat\Sigma_0^{-1/2}\), and set
\(\delta_j=u_j^\top\widehat\Sigma_0^{1/2}
(\widetilde\beta_0-\beta_0^\ast)\).

\begin{theorem}[New-task bias--variance decomposition]
\label{thm:new-task-transfer}
Under the conditions above,
\begin{equation}\label{eq:new-task-bias-variance}
\mathbb E\!\left[\frac1{n_0}\sum_{i=1}^{n_0}
\{\widehat f_0(X_{0i})-f_0^\ast(X_{0i})\}^2\right]
=\sum_{j=1}^d\left\{
\left(\frac{\lambda_0\gamma_j}{1+\lambda_0\gamma_j}\right)^2
\delta_j^2
+\frac{\sigma_0^2/n_0}{(1+\lambda_0\gamma_j)^2}
\right\},
\end{equation}
where \(f_0^\ast(x)=\widehat g(x)+\widehat z(x)^\top\beta_0^\ast\);
under the assumed exact model, this function equals
\(\mathbb E_0(Y\mid X=x)\). The expectation is over the new-task errors conditional on the original
fit and new-task covariates.
\end{theorem}

Equation~\eqref{eq:new-task-bias-variance} shows that directions with
\(\gamma_j=0\) receive no transfer. Larger \(\gamma_j\) produce greater
variance reduction, while their bias is governed by the discrepancy
between the new-task coefficient and the overlap-weighted center of the
original coefficients.

We next allow the learned functions to approximate, rather than exactly
span, the new-task conditional mean. Conditional on the original fit, set
\(\Sigma_0=\mathbb E_0\{\widehat z(X)\widehat z(X)^\top\}\) and define
\(\Omega_{0t}\) from \(\Sigma_0\) and \(\widehat\Sigma_t\) using
\eqref{eq:overlap}. For \(\|\beta\|_2\le B_\beta\), let
\[
\begin{aligned}
\mathcal R_0(\beta)
&=\frac12\mathbb E_0
\{Y-\widehat g(X)-\widehat z(X)^\top\beta\}^2,\\
\mathcal E_0(\beta)
&=\mathcal R_0(\beta)-\frac12\mathbb E_0\{Y-f_0^\ast(X)\}^2
=\frac12\mathbb E_0
\{\widehat g(X)+\widehat z(X)^\top\beta-f_0^\ast(X)\}^2,\\
\mathcal P_0(\beta)
&=\frac{\lambda_0}{2T}\sum_{t=1}^T
\|\beta-\widehat\beta_t\|_{\Omega_{0t}}^2.
\end{aligned}
\]
Here \(f_0^\ast(x)=\mathbb E_0(Y\mid X=x)\), so
\(\mathcal E_0\) is the excess prediction risk of the final new-task
predictor. Define the target second-moment rate
\[
r_{\Sigma,0}(\delta)=dB_z^2\left\{
\frac2{\sqrt{n_0}}+
\sqrt{\frac{2\log(d^2/\delta)}{n_0}}\right\},
\qquad
\Delta_0(\delta)=14\lambda_0B_\beta^2
\left(\frac{M_z}{m_z}\right)^2r_{\Sigma,0}(\delta).
\]

\begin{theorem}[New-task end-to-end oracle inequality]
\label{thm:new-task-oracle}
Let \(0<\delta<1/2\), and condition on an original fit satisfying the
bounds in Assumption~\ref{ass:neural-classes}, with \(|Y|\le B_Y\) under
the new task and
\(\max_{t\in[T]}\|\widehat\beta_t\|_2\le B_\beta\). Suppose, for the same
constants in Assumption~\ref{ass:uniform-spectrum}, that
\(m_zI_d\preceq\Sigma_0\preceq M_zI_d\) and
\(m_zI_d\preceq\widehat\Sigma_t\preceq M_zI_d\) for every \(t\in[T]\).
If \(r_{\Sigma,0}(\delta)\le m_z/2\), then, with conditional
probability at least \(1-2\delta\),
\begin{equation}\label{eq:new-task-oracle}
\begin{aligned}
\mathcal E_0(\widehat\beta_0)
\le\;&
\underbrace{
\inf_{\|\beta\|_2\le B_\beta}
\{\mathcal E_0(\beta)+\mathcal P_0(\beta)\}
}_{\text{target approximation and shrinkage}}
+\underbrace{
\frac{8BB_\beta B_z}{\sqrt{n_0}}
}_{\text{new-task coefficient estimation}}\\
&+\underbrace{
B^2\sqrt{\frac{\log(1/\delta)}{2n_0}}
}_{\text{concentration}}
+\underbrace{
2\Delta_0(\delta)
}_{\text{new-task overlap estimation}},
\end{aligned}
\end{equation}
where \(B=B_Y+B_g+B_zB_\beta\) as in
Section~\ref{sec:end-to-end-theory}.
\end{theorem}

The oracle term in \eqref{eq:new-task-oracle} includes the error incurred
when the learned \(\widehat g\) and \(\widehat z\) do not represent the
new-task conditional mean exactly. The final term is the additional cost
of learning the new-task overlap matrices. The spectral condition is used
only for this uniform overlap control; the adaptation rule and its range
safety continue to allow singular second-moment matrices. Thus the result
controls the final transferred predictor, while the preceding exact identity
isolates its direction-specific variance reduction.

The global bound can be sharpened when the penalized new-task oracle has
small risk. For the fixed affine predictor class used during adaptation,
define the localized scale
\[
\sqrt{r_0}
=\frac{(B_g+B_zB_\beta)\log^2(en_0)+B}{\sqrt{n_0}},
\]
and let
\(\beta_{0,\lambda_0}^\ast\in
\arg\min_{\|\beta\|_2\le B_\beta}
\{\mathcal R_0(\beta)+\mathcal P_0(\beta)\}\).
For brevity, set
\(s_0(\delta)=r_0+B^2\log(1/\delta)/n_0\).

\begin{theorem}[Optimistic new-task end-to-end oracle inequality]
\label{thm:new-task-optimistic}
Let \(0<\delta<1/3\), and suppose the conditions of
Theorem~\ref{thm:new-task-oracle} hold. Then a universal constant \(C>0\)
satisfies, with conditional probability at least \(1-3\delta\),
\begin{equation}\label{eq:new-task-optimistic}
\begin{aligned}
\mathcal E_0(\widehat\beta_0)
\le\;&
\underbrace{
\mathcal E_0(\beta_{0,\lambda_0}^\ast)
+\mathcal P_0(\beta_{0,\lambda_0}^\ast)
}_{\text{target approximation and shrinkage}}
+\underbrace{C\Delta_0(\delta)}_{\text{new-task overlap estimation}}\\
&+\underbrace{
C\left[
\sqrt{
\{\mathcal R_0(\beta_{0,\lambda_0}^\ast)
+\mathcal P_0(\beta_{0,\lambda_0}^\ast)+\Delta_0(\delta)\}
s_0(\delta)}
+s_0(\delta)
\right]
}_{\text{localized new-task estimation}}.
\end{aligned}
\end{equation}
\end{theorem}

Theorem~\ref{thm:new-task-optimistic} is the transfer counterpart of
Theorem~\ref{thm:optimistic-oracle}. When the penalized target oracle risk
and overlap-estimation error are both small, the localized prediction term
has the faster order \(s_0(\delta)\). The overlap term
remains explicit because estimating where the new task shares
representation support with the original tasks is part of the transfer
problem.

\subsection{\change{Proofs of the New-Task Results}}

\begin{proofof}{Theorem~\ref{thm:new-task-transfer}}
Let \(Z_0\) have rows \(\widehat z(X_{0i})^\top\), let
\(\varepsilon_0=(\varepsilon_{01},\ldots,\varepsilon_{0n_0})^\top\), and
set \(\xi_0=n_0^{-1}Z_0^\top\varepsilon_0\). Then
\(\widehat s_0=\widehat\Sigma_0\beta_0^\ast+\xi_0\),
\(\mathbb E(\xi_0)=0\), and
\(\operatorname{Cov}(\xi_0)=\sigma_0^2\widehat\Sigma_0/n_0\).
The normal equation for \eqref{eq:new-task-estimator}, followed by the
whitening transformation used in the proof of
Theorem~\ref{thm:fixed-bias-variance}, gives
\[
\widehat\Sigma_0^{1/2}(\widehat\beta_0-\beta_0^\ast)
=(I_d+\lambda_0M_0)^{-1}
\left\{\widehat\Sigma_0^{-1/2}\xi_0
+\lambda_0M_0\widehat\Sigma_0^{1/2}
(\widetilde\beta_0-\beta_0^\ast)\right\},
\]
where \(M_0=\widehat\Sigma_0^{-1/2}\widehat\Lambda_0
\widehat\Sigma_0^{-1/2}\). The transformed noise has covariance
\(\sigma_0^2I_d/n_0\). Expansion in the eigenbasis \(\{u_j\}_{j=1}^d\)
and conditional expectation give \eqref{eq:new-task-bias-variance}. The
left-hand side equals
\(\mathbb E\|\widehat\beta_0-\beta_0^\ast\|_{\widehat\Sigma_0}^2\),
and, as a prediction error, is unchanged by the preceding recentering.
\end{proofof}

\begin{proofof}{Theorem~\ref{thm:new-task-oracle}}
Condition on the original fit and apply the proof of
Theorem~\ref{thm:end-to-end} to the single-task affine class
\(x\mapsto\widehat g(x)+\widehat z(x)^\top\beta\). After subtracting the
loss at \(\beta=0\), symmetrization and contraction give a uniform risk
deviation bounded by
\[
4B\frac{B_\beta B_z}{\sqrt{n_0}}
+\frac{B^2}{2}\sqrt{\frac{\log(1/\delta)}{2n_0}}.
\]
Lemma~\ref{lem:second-moment}, specialized to the fixed singleton
\(\{\widehat z\}\) and one task, gives the stated rate because the
coordinate-product complexity of this singleton is at most
\(B_z^2/\sqrt{n_0}\). Hence
\(\|\widehat\Sigma_0-\Sigma_0\|_{\mathrm{op}}\le
r_{\Sigma,0}(\delta)\). Lemma~\ref{lem:overlap-stability}, with the
original matrices held fixed, then yields
\[
\max_{t\in[T]}\|\widehat\Omega_{0t}-\Omega_{0t}\|_{\mathrm{op}}
\le7\left(\frac{M_z}{m_z}\right)^2r_{\Sigma,0}(\delta).
\]
Since \(\|\beta-\widehat\beta_t\|_2\le2B_\beta\), the uniform difference
between the empirical and population transfer penalties is at most
\(\Delta_0(\delta)\). The same basic
inequality as in the proof of Theorem~\ref{thm:end-to-end} contributes
twice the risk and penalty deviations. Taking the infimum over \(\beta\)
therefore gives \eqref{eq:new-task-oracle}; a union bound gives probability
at least \(1-2\delta\).
\end{proofof}

\begin{proofof}{Theorem~\ref{thm:new-task-optimistic}}
Condition on the original fit. The Gaussian complexity of the affine
predictor class is at most
\((B_g+B_zB_\beta)/\sqrt{n_0}\), and its prediction diameter is at most
\(2B_zB_\beta\). The smooth-loss localization argument in the proof of
Proposition~\ref{prop:local-complexity} therefore gives the local
Rademacher majorant
\(C\{\sqrt{r_0\rho}+r_0\}\) at population risk level \(\rho\).
Consequently, the relative-deviation argument of
Lemma~\ref{lem:relative-deviation} applies with \(N\) and \(r_N\)
replaced by \(n_0\) and \(r_0\).

On the second-moment event used in the proof of
Theorem~\ref{thm:new-task-oracle}, the empirical transfer penalty differs
uniformly from \(\mathcal P_0\) by at most \(\Delta_0(\delta)\).
Bernstein's inequality at the fixed comparator
\(\beta_{0,\lambda_0}^\ast\), followed by the preceding relative-deviation
bound at \(\widehat\beta_0\), now gives exactly the three-event argument
in the proof of Theorem~\ref{thm:optimistic-oracle}. Substituting
\(n_0,r_0,\beta_{0,\lambda_0}^\ast\), and \(\Delta_0(\delta)\) for their
counterparts in that proof yields \eqref{eq:new-task-optimistic}. A union
bound over the penalty, comparator, and relative-deviation events gives
probability at least \(1-3\delta\).
\end{proofof}

\section{\change{Proofs for Sections~\ref{sec:method} and \ref{sec:theory}}}
\label{app:main-proofs}

This appendix proves the results in Sections~\ref{sec:method} and \ref{sec:theory} in their order of appearance.

\subsection{\change{Proofs for Section~\ref{sec:method}}}\label{app:method-proofs}

\begin{proofof}{Proposition~\ref{prop:pooling}}
Let \(A=\Sigma_t\), \(B=\Sigma_s\), and \(\delta=\beta_t-\beta_s\). With \(u=\beta_t-a\), the criterion in \eqref{eq:pooling-cost} becomes
\[
q(u)=u^\top A u+(u-\delta)^\top B(u-\delta).
\]
For \(A,B\succeq0\), \(\Null(A+B)=\Null(A)\cap\Null(B)\), so \(\Range(B)\subseteq\Range(A+B)\). The normal equation \((A+B)u=B\delta\) is therefore solvable, with one solution \(u_0=(A+B)^\dagger B\delta\). Substitution gives
\[
\begin{aligned}
\mathcal C_{ts}(\beta_t,\beta_s)
&=u_0^\top(A+B)u_0-2\delta^\top Bu_0+\delta^\top B\delta\\
&=\delta^\top\{B-B(A+B)^\dagger B\}\delta\\
&=\delta^\top A(A+B)^\dagger B\delta
=\frac12\delta^\top\Omega_{ts}\delta.
\end{aligned}
\]
The second equality uses
\((A+B)u_0=B\delta\). For the third,
\((A+B)(A+B)^\dagger B=B\) since
\(\Range(B)\subseteq\Range(A+B)\); subtracting
\(B(A+B)^\dagger B\) from both sides gives
\(A(A+B)^\dagger B=B-B(A+B)^\dagger B\).
The matrix \(B-B(A+B)^\dagger B\) is symmetric, and its quadratic form is nonnegative as it is the minimum of \(q\). Hence \(\Omega_{ts}\) is symmetric and positive semidefinite, and the identity in \eqref{eq:overlap} follows. To characterize its null space, observe that the minimum is zero if and only if there is an \(a\) such that
\(\beta_t-a\in\Null(A)\) and
\(\beta_s-a\in\Null(B)\). Subtracting these two relations shows that
\(\delta\in\Null(A)+\Null(B)\), and the converse follows by reversing the same construction. Therefore
\[
\Null(\Omega_{ts})=\Null(A)+\Null(B),
\qquad
\Range(\Omega_{ts})=\Range(A)\cap\Range(B),
\]
where the second equality follows by taking orthogonal complements. Finally, exchanging \(A\) and \(B\) leaves the pooling problem unchanged. Returning to the original variable \(a=\beta_t-u\), the normal equation is
\((A+B)a=A\beta_t+B\beta_s\). Thus all minimizing auxiliary coefficients are
\(a_{ts}^{\ast}=(A+B)^\dagger(A\beta_t+B\beta_s)+v\), with \(v\in\Null(A+B)\).

For completeness, when \(A\) and \(B\) are positive definite, the matrix
\(A(A+B)^{-1}B\) is invertible and
\[
\{A(A+B)^{-1}B\}^{-1}
=B^{-1}(A+B)A^{-1}=A^{-1}+B^{-1}.
\]
Consequently, \(\Omega_{ts}=2(A^{-1}+B^{-1})^{-1}\). This also makes the symmetry of the positive-definite expression explicit.
\end{proofof}

\begin{proofof}{Corollary~\ref{cor:penalty-comparison}}
Let \(A=\Sigma_t\), \(B=\Sigma_s\), and \(S=A+B\). Since \(A\) and \(B\) are positive semidefinite, their ranges are contained in \(\Range(S)\). Hence \(AS^\dagger S=A\) and \(BS^\dagger S=B\), which give
\[
AS^\dagger A=A-AS^\dagger B,
\qquad
BS^\dagger B=B-BS^\dagger A.
\]
The pairwise pooling identity implies that \(AS^\dagger B=BS^\dagger A=\Omega_{ts}/2\). Expanding the right-hand side of \eqref{eq:penalty-comparison} therefore yields
\[
\begin{aligned}
\frac12(A-B)S^\dagger(A-B)
&=\frac12\{AS^\dagger A-AS^\dagger B-BS^\dagger A+BS^\dagger B\}\\
&=\frac{A+B}{2}-2AS^\dagger B
=\frac{\Sigma_t+\Sigma_s}{2}-\Omega_{ts}.
\end{aligned}
\]
The matrix on the first line is positive semidefinite since \(S^\dagger\succeq0\), proving \eqref{eq:penalty-comparison}. Applying the resulting Loewner inequality to \(\beta_t-\beta_s\) gives the comparison of the two quadratic penalties. If \(\Sigma_t=\Sigma_s\), the correction vanishes and the matrices are equal. The same conclusion also follows directly from \eqref{eq:overlap} and the identities \((c\Sigma)^\dagger=c^{-1}\Sigma^\dagger\) for \(c>0\) and \(\Sigma\Sigma^\dagger\Sigma=\Sigma\): when \(\Sigma_t=\Sigma_s=\Sigma\), \(\Omega_{ts}=2\Sigma(2\Sigma)^\dagger\Sigma=\Sigma\). This remains valid when \(\Sigma\) is singular.

When every task has this same second-moment matrix, expand the complete pairwise sum over ordered task pairs and divide by two:
\[
\begin{aligned}
\sum_{1\le t<s\le T}\|\beta_t-\beta_s\|_\Sigma^2
&=
\frac12\sum_{t=1}^T\sum_{s=1}^T
(\beta_t-\beta_s)^\top\Sigma(\beta_t-\beta_s)\\
&=
T\sum_{t=1}^T\|\beta_t\|_\Sigma^2
-
\left\|\sum_{t=1}^T\beta_t\right\|_\Sigma^2.
\end{aligned}
\]
To obtain the last equality, the two quadratic terms contribute \(T\sum_{t=1}^T\beta_t^\top\Sigma\beta_t\), while the cross terms equal \(-(\sum_{t=1}^T\beta_t)^\top\Sigma(\sum_{s=1}^T\beta_s)\). The latter vanishes under the centering constraint, which proves the result.
\end{proofof}

\subsection{\change{Proofs for Section~\ref{sec:theory}}}\label{app:theory-proofs}

\begin{proofof}{Proposition~\ref{prop:pure-covariate}}
For task \(t\), write \(Y=g^\ast(X)+\varepsilon_t\), where \(\mathbb E_t(\varepsilon_t\mid X)=0\). Expanding the squared loss for an arbitrary \(\beta_t\) yields a noise term, a squared prediction-deviation term, and the cross term \(-\mathbb E_t\{\varepsilon_t z(X)^\top\beta_t\}\). The cross term is zero since
\(\mathbb E_t[\varepsilon_t z(X)^\top\beta_t]=\mathbb E_t[\mathbb E_t(\varepsilon_t\mid X)z(X)^\top\beta_t]=0\). Consequently,
\[
\begin{aligned}
&\frac12\mathbb E_t\{Y-g^\ast(X)-z(X)^\top\beta_t\}^2
-
\frac12\mathbb E_t\{Y-g^\ast(X)\}^2\\
&\hspace{4cm}
=
\frac12\mathbb E_t\{z(X)^\top\beta_t\}^2
=
\frac12\|\beta_t\|_{\Sigma_t}^2.
\end{aligned}
\]
Subtracting the population objective at the feasible point \(\beta_1=\cdots=\beta_T=0\) therefore gives
\[
\frac1{2T}\sum_{t=1}^T\|\beta_t\|_{\Sigma_t}^2
+\frac{\lambda}{T(T-1)}
\sum_{1\le t<s\le T}\|\beta_t-\beta_s\|_{\Omega_{ts}}^2.
\]
Both terms are nonnegative, so the zero coefficients are population minimizers. If every \(\Sigma_t\succ0\), the first term is zero only if \(\beta_t=0\) for every task; hence the minimizer is unique. If some \(\Sigma_t\) is singular, any minimizer must still make every summand in the first term zero. Since \(\|\beta_t\|_{\Sigma_t}^2=\mathbb E_t\{z(X)^\top\beta_t\}^2\), this is equivalent to \(z(X)^\top\beta_t=0\) almost surely under \(\mathbb P_t^X\). Thus covariate heterogeneity can change the second-moment matrices, but under correct specification it cannot by itself create a nonzero task-specific conditional-mean function.
\end{proofof}

\begin{proofof}{Proposition~\ref{prop:invariance}}
The invertibility of \(L\) makes \(\beta_{t,L}=L^{-\top}\beta_t\) a one-to-one reparameterization of the coefficient vector. Direct substitution gives the prediction and second-moment identities
\[
(Lz)^\top L^{-\top}\beta_t=z^\top\beta_t,
\qquad
\mathbb E_t\{Lz(X)z(X)^\top L^\top\}=L\Sigma_tL^\top.
\]
The centering constraint is also preserved since \(\sum_{t=1}^T\beta_{t,L}=L^{-\top}\sum_{t=1}^T\beta_t=0\).

It remains to verify the transformation of the covariate-overlap penalty, for which the variational definition is more useful than manipulating the Moore--Penrose inverse directly. For any auxiliary coefficient \(a\), set \(a_L=L^{-\top}a\). Then
\[
\|\beta_{t,L}-a_L\|_{\Sigma_{t,L}}^2
=
\|\beta_t-a\|_{\Sigma_t}^2,
\]
and the same calculation holds for task \(s\). Since \(a\mapsto a_L\) is bijective, minimizing over the original and transformed auxiliary coefficients gives the same pooling cost. Let \(\delta=\beta_t-\beta_s\) and \(\delta_L=L^{-\top}\delta\). Equation~\eqref{eq:overlap} then yields
\[
\frac12\delta_L^\top\Omega_{ts,L}\delta_L
=\mathcal C_{ts,L}
=\mathcal C_{ts}
=\frac12\delta_L^\top L\Omega_{ts}L^\top\delta_L.
\]
Every \(\delta_L\in\mathbb R^d\) can arise from some \(\delta\) since \(L\) is invertible. Both matrices in the two quadratic forms are symmetric, so equality of their quadratic forms for every \(\delta_L\) implies \(\Omega_{ts,L}=L\Omega_{ts}L^\top\). Substituting this identity into the quadratic penalty proves its invariance and completes the proof.
\end{proofof}

\begin{proofof}{Theorem~\ref{thm:fixed-bias-variance}}
Write \(Z_t\) for the \(n\times d\) matrix whose \(i\)th row is \(z(X_{ti})^\top\), and let \(\varepsilon_t=(\varepsilon_{t1},\ldots,\varepsilon_{tn})^\top\). Define
\(\xi_t=n^{-1}Z_t^\top\varepsilon_t\) and stack the resulting vectors as
\(\xi=(\xi_1^\top,\ldots,\xi_T^\top)^\top\). Conditional on \(\mathscr X\), Assumption~\ref{ass:coefficient-model} gives
\[
\mathbb E(\xi_t\mid\mathscr X)=0,
\qquad
\operatorname{Cov}(\xi_t\mid\mathscr X)
=\frac{\sigma^2}{n^2}Z_t^\top Z_t
=\frac{\sigma^2}{n}\widehat\Sigma_t.
\]
The errors are conditionally independent across tasks, so the off-diagonal covariance blocks vanish. It follows that
\(\mathbb E(\xi\mid\mathscr X)=0\) and
\(\operatorname{Cov}(\xi\mid\mathscr X)=(\sigma^2/n)\widehat D\).

We next express the empirical criterion as a quadratic function of the centered coefficient vector. Under the model in Assumption~\ref{ass:coefficient-model}, the residual vector in task \(t\) is
\(\varepsilon_t-Z_t(\beta_t-\beta_t^\ast)\). Expanding its squared norm, summing over tasks, and multiplying \eqref{eq:estimator} by \(T\) leaves, up to a term independent of \(\beta\),
\[
Q(\beta)
=\frac12(\beta-\beta^\ast)^\top\widehat D(\beta-\beta^\ast)
-\xi^\top(\beta-\beta^\ast)
+\frac{\kappa}{2}\beta^\top\widehat L\beta,
\]
where \(\kappa=2\lambda/(T-1)\). The coefficient of the last term follows from
\(\lambda(T-1)^{-1}\beta^\top\widehat L\beta
=(\kappa/2)\beta^\top\widehat L\beta\).

The feasible set is the centered subspace. Therefore the first-order condition at \(\widehat\beta\) is
\[
u^\top\{(\widehat D+\kappa\widehat L)\widehat\beta
-\widehat D\beta^\ast-\xi\}=0
\quad\text{for every centered }u.
\]
Assumption~\ref{ass:moment-condition} makes \(\widehat D\) positive definite on this subspace. Since \(\widehat L\succeq0\), the Hessian
\(\widehat D+\kappa\widehat L\) is also positive definite there, and the first-order condition identifies the unique centered minimizer.

We briefly verify the existence of the generalized eigenbasis used in
\eqref{eq:spectral}. Let \(C\) be any \(Td\times(T-1)d\) matrix whose columns form a Euclidean orthonormal basis of the centered subspace. Assumption~\ref{ass:moment-condition} implies
\(C^\top\widehat DC\succ0\). Therefore
\[
(C^\top\widehat DC)^{-1/2}
C^\top\widehat LC
(C^\top\widehat DC)^{-1/2}
\]
is symmetric and positive semidefinite. Let \(w_j\) be its Euclidean orthonormal eigenvectors, with eigenvalues \(\mu_j\), and set
\(v_j=C(C^\top\widehat DC)^{-1/2}w_j\).
Direct multiplication shows that
\(v_j^\top\widehat Dv_k=\delta_{jk}\) and that these vectors satisfy
\eqref{eq:spectral}. This construction also shows \(\mu_j\ge0\).

Take \(u=v_j\) in the first-order condition. By symmetry of \(\widehat L\) and the generalized eigenrelation \eqref{eq:spectral},
\[
v_j^\top\widehat L\widehat\beta
=\widehat\beta^\top\widehat Lv_j
=\mu_j\widehat\beta^\top\widehat Dv_j
=\mu_jv_j^\top\widehat D\widehat\beta.
\]
Let \(\widehat\alpha_j=v_j^\top\widehat D\widehat\beta\) and
\(\zeta_j=v_j^\top\xi\). Because
\(\theta_j=v_j^\top\widehat D\beta^\ast\), the projected normal equation becomes
\((1+\kappa\mu_j)\widehat\alpha_j=\theta_j+\zeta_j\), and hence
\[
\widehat\alpha_j-\theta_j
=-\frac{\kappa\mu_j}{1+\kappa\mu_j}\theta_j
+\frac{\zeta_j}{1+\kappa\mu_j}.
\]

The conditional covariance of the projected noise coordinates is
\[
\operatorname{Cov}(\zeta_j,\zeta_k\mid\mathscr X)
=v_j^\top\operatorname{Cov}(\xi\mid\mathscr X)v_k
=\frac{\sigma^2}{n}v_j^\top\widehat Dv_k
=\frac{\sigma^2}{n}\delta_{jk}.
\]
In particular, each \(\zeta_j\) has conditional mean zero and conditional variance \(\sigma^2/n\).

Both \(\widehat\beta\) and \(\beta^\ast\) are centered. Since the generalized eigenvectors form a \(\widehat D\)-orthonormal basis of the centered subspace,
\[
\widehat\beta-\beta^\ast
=\sum_{j=1}^{(T-1)d}
(\widehat\alpha_j-\theta_j)v_j,
\qquad
\|\widehat\beta-\beta^\ast\|_{\widehat D}^2
=\sum_{j=1}^{(T-1)d}
(\widehat\alpha_j-\theta_j)^2.
\]
The left-hand side equals
\(\sum_{t=1}^T\|\widehat\beta_t-\beta_t^\ast\|_{\widehat\Sigma_t}^2\).
Squaring the coordinate representation, taking its conditional expectation, and using the zero conditional mean of \(\zeta_j\) removes the bias--noise cross term. The squared deterministic component and the noise variance are respectively
\[
\left(\frac{\kappa\mu_j}{1+\kappa\mu_j}\right)^2\theta_j^2
\quad\text{and}\quad
\frac{\sigma^2/n}{(1+\kappa\mu_j)^2}.
\]
Summing these terms over the centered coordinates and dividing by \(T\) proves \eqref{eq:fixed-bias-variance}.
\end{proofof}

\begin{proofof}{Corollary~\ref{cor:equal-moment-shrinkage}}
When the empirical second-moment matrices are equal, \eqref{eq:overlap} gives
\(\widehat\Omega_{ts}=\widehat\Sigma\), while
\(\widehat D=I_T\otimes\widehat\Sigma\). Expanding the sum of pairwise quadratic forms shows that the matrix-weighted complete-graph Laplacian is
\[
\widehat L=(TI_T-1_T1_T^\top)\otimes\widehat\Sigma,
\]
where \(1_T\in\mathbb R^T\) is the vector of ones. On the centered task subspace, \((TI_T-1_T1_T^\top)\) acts as \(TI_T\). Hence \(\widehat Lv=T\widehat Dv\) for every centered \(v\), so all the generalized eigenvalues in \eqref{eq:spectral} equal \(T\). Substitution into Theorem~\ref{thm:fixed-bias-variance} gives the stated denominator.
\end{proofof}

\begin{proofof}{Theorem~\ref{thm:end-to-end}}
We first control the empirical risk uniformly over the jointly learned functions. Assumption~\ref{ass:neural-classes} and the coefficient bound imply
\(|f_{h,t}(x)|\le B_g+B_zB_\beta\) and
\(|Y-f_{h,t}(X)|\le B\). Set
\[
\psi_{h,t}(x,y)
=\ell_{h,t}(x,y)-\frac12y^2,
\qquad
U_{\psi}
=
\sup_{h\in\mathcal H}
\left|
\frac1N\sum_{t=1}^T\sum_{i=1}^n\psi_{h,t}(X_{ti},Y_{ti})
-\frac1T\sum_{t=1}^T\mathbb E_t\psi_{h,t}(X,Y)
\right|.
\]
Let \(\{(X_{ti}',Y_{ti}')\}_{t\in[T],i\in[n]}\) be an independent ghost sample, with each observation drawn from the same task as its unprimed counterpart. By Jensen's inequality and symmetrization,
\[
\begin{aligned}
\mathbb E U_{\psi}
&\le
\mathbb E\sup_{h\in\mathcal H}
\left|
\frac1N\sum_{t=1}^T\sum_{i=1}^n
\{\psi_{h,t}(X_{ti},Y_{ti})
-\psi_{h,t}(X_{ti}',Y_{ti}')\}
\right|\\
&\le
2\mathbb E\sup_{h\in\mathcal H}
\left|
\frac1N\sum_{t=1}^T\sum_{i=1}^n
\epsilon_{ti}\psi_{h,t}(X_{ti},Y_{ti})
\right|.
\end{aligned}
\]
For fixed \(y\), the map
\(u\mapsto\{(y-u)^2-y^2\}/2\) vanishes at zero and is
\(B\)-Lipschitz on the range of the predictors. Because our Rademacher complexity is defined with an absolute value, the two-sided contraction inequality gives
\[
\mathbb E U_{\psi}
\le4B\,\mathscr R_N(\mathcal F).
\]
Elementary maximization over \(|y|\le B_Y\) and
\(|u|\le B_g+B_zB_\beta\) shows that the range length of
\(\{(y-u)^2-y^2\}/2\) is at most
\((B_Y+B_g+B_zB_\beta)^2/2=B^2/2\).
Replacing one of the \(N\) observations can therefore change
\(U_{\psi}\) by at most \(B^2/(2N)\). McDiarmid's inequality gives
\[
\Pr\{U_{\psi}-\mathbb EU_{\psi}>u\}
\le
\exp\left\{-\frac{8Tn\,u^2}{B^4}\right\}.
\]
Taking \(u=(B^2/2)\{\log(1/\delta)/(2Tn)\}^{1/2}\) shows that, with probability at least \(1-\delta\),
\[
U_{\psi}
\le
4B\,\mathscr R_N(\mathcal F)
+\frac{B^2}{2}
\sqrt{\frac{\log(1/\delta)}{2Tn}}.
\]
The observations need not be identically distributed for these arguments: independence is sufficient, and each empirical term is paired with the expectation under the same task distribution.

We next control the random penalty. By Lemma~\ref{lem:second-moment}, an event of probability at least \(1-\delta\) satisfies
\(\Delta_\Sigma\le r_{\Sigma,n}(\delta)\).
The assumption \(r_{\Sigma,n}(\delta)\le m_z/2\) allows Lemma~\ref{lem:overlap-stability} to be applied on this event. Write
\[
\Delta_\Omega
=\sup_{z\in\mathcal Z}\max_{1\le t<s\le T}
\|\widehat\Omega_{ts}(z)-\Omega_{ts}(z)\|_{\mathrm{op}}.
\]
For any symmetric matrix \(A\) and vector \(v\),
\(|v^\top Av|\le\|A\|_{\mathrm{op}}\|v\|_2^2\).
Since \(\|\beta_t-\beta_s\|_2\le2B_\beta\), each covariate-overlap penalty difference is at most
\(4B_\beta^2\Delta_\Omega\). There are \(T(T-1)/2\) unordered task pairs, and hence
\[
\begin{aligned}
U_{\mathcal P}
&:=
\sup_{h\in\mathcal H}
|\widehat{\mathcal P}_\lambda(h)-\mathcal P_\lambda(h)|\\
&\le
\frac{\lambda}{T(T-1)}
\sum_{1\le t<s\le T}4B_\beta^2\Delta_\Omega
=2\lambda B_\beta^2\Delta_\Omega\\
&\le
14\lambda B_\beta^2
\left(\frac{M_z}{m_z}\right)^2
r_{\Sigma,n}(\delta).
\end{aligned}
\]

It remains to combine these two uniform bounds with empirical optimality. Set
\[
c=T^{-1}\sum_{t=1}^T\mathbb E_t(Y^2/2)
-N^{-1}\sum_{t=1}^T\sum_{i=1}^nY_{ti}^2/2.
\]
For every \(h\in\mathcal H\),
\[
\mathcal R(h)-\widehat{\mathcal R}(h)
=T^{-1}\sum_{t=1}^T\mathbb E_t\psi_{h,t}
-N^{-1}\sum_{t=1}^T\sum_{i=1}^n\psi_{h,t}(X_{ti},Y_{ti})+c.
\]
Consequently,
\[
\begin{aligned}
\mathcal R(\widehat h)+\mathcal P_\lambda(\widehat h)
&\le
\widehat{\mathcal R}(\widehat h)
+\widehat{\mathcal P}_\lambda(\widehat h)
+U_{\psi}+U_{\mathcal P}+c\\
&\le
\widehat{\mathcal R}(h)
+\widehat{\mathcal P}_\lambda(h)
+U_{\psi}+U_{\mathcal P}+c\\
&\le
\mathcal R(h)+\mathcal P_\lambda(h)
+2U_{\psi}+2U_{\mathcal P}.
\end{aligned}
\]
The common \(Y^2/2\) term cancels in the last step. The first and third inequalities are uniform, and the middle inequality is the definition of \(\widehat h\). Because
\(\mathcal P_\lambda(\widehat h)\ge0\), it may be removed from the left-hand side. On the intersection of the two concentration events,
\[
\begin{aligned}
2U_{\psi}
&\le
8B\,\mathscr R_N(\mathcal F)
+B^2\sqrt{\frac{\log(1/\delta)}{2Tn}},\\
2U_{\mathcal P}
&\le
28\lambda B_\beta^2
\left(\frac{M_z}{m_z}\right)^2
r_{\Sigma,n}(\delta).
\end{aligned}
\]
Subtracting \(\mathcal R(f^\ast)\) turns the population-risk difference into \(\mathcal E\). The preceding basic inequality holds for every
\(h\in\mathcal H\), so taking its infimum and substituting these last two bounds proves \eqref{eq:oracle}. A union bound shows that the two events hold simultaneously with probability at least \(1-2\delta\).
\end{proofof}

\begin{proofof}{Theorem~\ref{thm:optimistic-oracle}}
If the population minimum is not attained, the argument applies to an
approximating sequence and its approximation error can then be sent to
zero. For the proof, set
\[
R_\lambda^\ast
=\mathcal R(h_\lambda^\ast)+\mathcal P_\lambda(h_\lambda^\ast),
\qquad
b=\frac{B^2}{2}.
\]
Write \(x=\log(1/\delta)\) and
\[
s=r_N+\frac{B^2x}{N}.
\]

We work on the intersection of three events. First,
Lemmas~\ref{lem:second-moment} and
\ref{lem:overlap-stability}, together with the calculation in the
proof of Theorem~\ref{thm:end-to-end}, imply with probability at least
\(1-\delta\) that
\[
\sup_{h\in\mathcal H}
\left|
\widehat{\mathcal P}_\lambda(h)
-\mathcal P_\lambda(h)
\right|
\le\Delta_{\lambda,n}(\delta).
\]
Second, the \(N\) loss variables evaluated at the fixed comparator
\(h_\lambda^\ast\) are independent, lie in \([0,b]\), and satisfy
\[
\frac1T\sum_{t=1}^T
\mathbb E_t\ell_{h_\lambda^\ast,t}^2
\le
b\mathcal R(h_\lambda^\ast).
\]
Bernstein's inequality therefore gives, with probability at least
\(1-\delta\),
\[
\widehat{\mathcal R}(h_\lambda^\ast)
\le
\mathcal R(h_\lambda^\ast)
+\sqrt{\frac{2b\mathcal R(h_\lambda^\ast)x}{N}}
+\frac{2bx}{3N}.
\]
Third, Lemma~\ref{lem:relative-deviation} holds with probability at
least \(1-\delta\).

On the intersection of these events, empirical optimality and
nonnegativity of the empirical penalty give
\[
\begin{aligned}
\widehat{\mathcal R}(\widehat h)
&\le
\widehat{\mathcal R}(\widehat h)
+\widehat{\mathcal P}_\lambda(\widehat h)\\
&\le
\widehat{\mathcal R}(h_\lambda^\ast)
+\widehat{\mathcal P}_\lambda(h_\lambda^\ast)\\
&\le
R_\lambda^\ast+\Delta_{\lambda,n}(\delta)
+\sqrt{\frac{2bR_\lambda^\ast x}{N}}
+\frac{2bx}{3N}.
\end{aligned}
\]
The last step uses
\(\mathcal R(h_\lambda^\ast)\le R_\lambda^\ast\).
Because \(bx/N\le s\), the right-hand side is at most
\[
Q
=R_\lambda^\ast+\Delta_{\lambda,n}(\delta)
+C\sqrt{R_\lambda^\ast s}+Cs.
\]

Apply Lemma~\ref{lem:relative-deviation} to \(\widehat h\). Writing
\(u=K-1\ge1\), we obtain
\[
\mathcal R(\widehat h)
\le
Q+\frac Qu+C(1+u)s.
\]
If \(s>0\), choosing \(u=\max\{1,\sqrt{Q/(Cs)}\}\), so that the
corresponding \(K=1+u\) is at least two, gives
\[
\mathcal R(\widehat h)
\le Q+Cs+C\sqrt{Qs}.
\]
The definition of \(Q\), followed twice by
\(\sqrt{a+c\sqrt{as}+cs}\le C(\sqrt a+\sqrt s)\), yields
\[
\mathcal R(\widehat h)
\le
R_\lambda^\ast
+C\left[
\Delta_{\lambda,n}(\delta)
+\sqrt{\{R_\lambda^\ast+\Delta_{\lambda,n}(\delta)\}s}
+s
\right].
\]
The case \(s=0\) follows by continuity. Subtracting
\(\mathcal R(f^\ast)\), using
\(\mathcal E(\widehat h)
=\mathcal R(\widehat h)-\mathcal R(f^\ast)\), proves
\eqref{eq:optimistic-oracle}, since
\(R_\lambda^\ast-\mathcal R(f^\ast)
=\mathcal E(h_\lambda^\ast)+\mathcal P_\lambda(h_\lambda^\ast)\).
A union bound over the three events gives
probability at least \(1-3\delta\).
\end{proofof}

\bibliography{sample}

\end{document}